\documentclass{article} 

\usepackage[final]{colm2026_conference}

\usepackage{microtype}
\usepackage{hyperref}
\usepackage{url}
\usepackage{booktabs}

\usepackage{lineno}

\usepackage{graphicx}
\usepackage{booktabs}
\usepackage[table]{xcolor} 
\usepackage{amsmath,amssymb,bm}
\usepackage{multirow}
\usepackage{xcolor}
\usepackage[table]{xcolor}

\usepackage{graphicx}
\usepackage{booktabs}
\usepackage[table]{xcolor} 
\usepackage{caption}
\usepackage{booktabs}
\usepackage{subcaption}   
\usepackage{amssymb}      
\usepackage{algorithm}            
\usepackage{algorithmicx}         
\usepackage{algpseudocode}        
\usepackage[accsupp]{axessibility}  

\definecolor{darkblue}{rgb}{0, 0, 0.5}
\hypersetup{colorlinks=true, citecolor=darkblue, linkcolor=darkblue, urlcolor=darkblue}

\title{DCPI: Dataset Condensation using Privileged Information} 

\author{{
    \bf Shaobo Wang$^{1}$\quad Youxin Jiang$^{1}$ \quad Tianle Niu$^{1}$ \quad Yantai Yang$^{1}$ \quad Ruiji Zhang$^{1}$
    \vspace{3pt}
  } \\
  {
  \bf Shuhao Hu$^{1}$ \quad Shuaiyu Zhang$^{1}$ \quad Chenghao Sun$^{1}$ \quad Weiya Li$^{2}$ \quad Conghui He$^{3}$
    \vspace{3pt}
  } \\
  {
  \bf Xuming Hu$^{4}$ \quad Linfeng Zhang$^{1}$\thanks{Corresponding author.}
    \vspace{8pt}
  } \\
  {
  $^1$ EPIC Lab, SJTU \quad $^2$ ICBC \quad $^3$ Shanghai AI Lab \quad $^4$ HKUST(GZ)
  \vspace{3pt}
  }
}

\begin{document}

\ifcolmsubmission
\linenumbers
\fi

\maketitle

\begin{abstract}

Dataset Condensation (DC) seeks to select or distill samples from large datasets into smaller subsets while preserving performance on target tasks. Existing methods primarily focus on pruning or synthesizing data in the same format as the original dataset, typically being the input data and corresponding labels. However, in DC settings, we find it is possible to synthesize more information beyond the data-label pair as an additional learning target to facilitate model training. In this paper, we introduce Dataset Condensation using Privileged Information (DCPI), which enriches DC by synthesizing privileged information alongside the reduced dataset. This privileged information can take the form of feature labels or attention labels, providing auxiliary supervision to improve model learning. Our findings reveal that effective feature labels must balance between being overly discriminative and excessively diverse, with a moderate level proves optimal for improving the reduced dataset’s efficacy. Extensive experiments on ImageNet-1K, CIFAR-10/100 and Tiny ImageNet demonstrate that DCPI integrates seamlessly with existing dataset condensation methods, offering significant performance gains.

\end{abstract}    
\section{Introduction}
\label{sec:intro}

Dataset Condensation (DC) has attracted considerable attention in recent years, with the primary aim of compressing large datasets into smaller subsets while maintaining comparable statistical performance. The existing DC methods can be broadly classified into two main categories: \textit{coreset selection} and \textit{dataset distillation}. Coreset selection methods focus on selecting a subset of samples from the original dataset~\cite{k_center, herding, forgetting}, while dataset distillation involves synthesizing unseen samples from the dataset~\cite{DD, DC, DM, MTT, SDC}.

\begin{figure}[tb!]
    \centering
    \vspace{-5pt}
\includegraphics[width=0.99\linewidth]{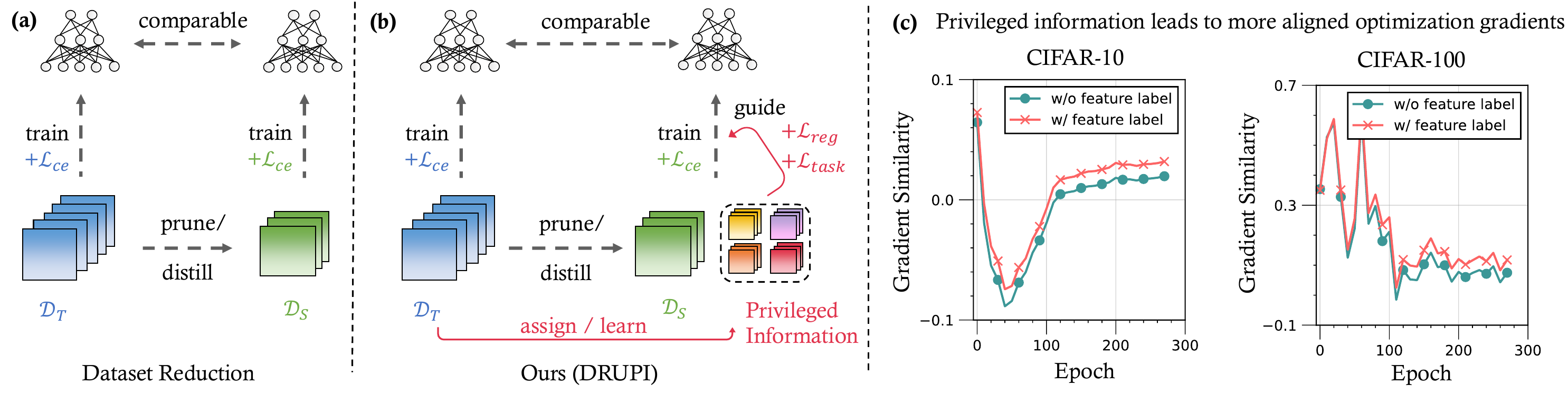}
    \vspace{-5pt}
    \caption{A comparison between conventional dataset reduction pipelines and our proposed DCPI framework.
    (a) Previous dataset reduction methods distill or select a subset $\mathcal{D}_{\mathcal{S}}$ from the original dataset $\mathcal{D}_{\mathcal{T}}$, maintaining the original ``data-label’’ structure.
    (b) In contrast, DCPI synthesizes auxiliary privileged information from $\mathcal{D}_{\mathcal{T}}$, enriching further supervision to models trained on the reduced subset $\mathcal{D}_{\mathcal{S}}$. (c) Cosine similarity between the gradients of a pre-trained model on synthetic datasets w/ and w/o privileged information (feature labels) and the real dataset. Synthetic datasets are generated using DC with 10 images per class (IPC) on the CIFAR-10 and CIFAR-100 datasets. We used the same pre-trained ConvNet for gradient extraction.}
    \label{fig:pipeline}
    \vspace{-17pt}
\end{figure}

In typical real-world scenarios, training of models for target tasks are generally restricted to input data (\emph{e.g.}, images) and their corresponding labels, as these are the most readily available elements. Although existing DC methods have shown strong performance~\cite{DD, DC, DM, MTT, sre2l}, they typically do so by compressing datasets in the same or similar format, such as the conventional data-label structure. Even advanced dataset distillation techniques, which re-parameterize images or labels to create alternative representations~\cite{IDC, IDM, Linba, Haba, speed}, are restricted by this conventional framework. As illustrated in Figure~\ref{fig:pipeline}(a), this reliance on fixed data-label structures restricts the ability of such methods to incorporate richer information that could further enhance model training and improve generalization.

In fact, DC settings offer the potential to create more diverse compressed datasets that extend beyond the simple input data $x_i$ and labels $y_i$, incorporating richer forms of information. A notable example is the concept of \textit{privileged information}, first introduced in the context of statistical learning~\cite{vapnik2009new,privileged_info,wang2026grounding}. Figure~\ref{fig:PI_format} provides an illustration of the privileged information. Let us consider a more concrete example, where $x_i$ might represent a biopsy image, and the privileged information $f_i^\star$ for $x_i$ could be the oncologist’s written assessment of the image~\cite{vapnik2015learning}. The label $y_i$ would then indicate whether the tissue in the image is malignant or benign. By leveraging this privileged information $f_i^\star$, a medical expert can make more informed decisions, benefiting from additional insights that improve diagnostic accuracy.

However, none of the existing methods compress the original dataset beyond the traditional data-label structure or synthesize privileged information for auxiliary supervision. To address this gap, we introduce a novel approach that, for the first time, \textbf{incorporates not only images and labels but also privileged information}. Our method, called \underline{\textbf{D}}ataset \underline{\textbf{C}}ondensation using \underline{\textbf{P}}rivileged \underline{\textbf{I}}nformation (DCPI), is illustrated in Figure~\ref{fig:pipeline}(b). We primarily synthesize feature labels for the reduced dataset, as these labels capture richer, high-dimensional information, enhancing dataset quality. These feature labels generalize effectively across various neural network architectures and provide a unified representation of latent statistics across multiple models, offering additional supervision during training. Additionally, we propose a more efficient form of feature labels, \emph{i.e.}, attention labels. As shown in Figure~\ref{fig:pipeline}(c), the incorporation of privileged information produces gradients more aligned with those of the original dataset, ultimately improving the generalization capabilities of the model.

A key challenge for DCPI lies in determining the appropriate feature labels to synthesize for the reduced dataset. To address this, we employ an arbitrary method of dataset distillation to synthesize the feature labels. During each step of the bi-level optimization, we match the statistical information of models trained on reduced datasets with and without feature labels. Furthermore, we observed that synthesized feature labels cannot be overly discriminative or diverse, which degrades the overall quality of the reduced dataset. This finding suggests that an optimal balance between discriminability and diversity is crucial for synthesizing effective feature labels. Our contributions are summarized as follows:
\begin{enumerate}
\item We propose a new paradigm, \emph{i.e.}, DCPI, for dataset reduction. In particular, privileged information, such as feature labels, can be synthesized in addition to traditional data-label pairs. This privileged information provides additional generative supervision during model training, thereby improving the generalization ability of the reduced dataset.
\item We observe that effective feature labels should balance the trade-off between diversity and discriminability. Overly discriminative feature labels, such as those directly extracted from a pre-trained neural network, can even degrade the quality of the reduced dataset.
\item We further provide a theoretical analysis of our DCPI pipeline based on VC theory~\cite{vc_theory} from the statistical learning, which rigorously ensures its effectiveness.
\item Our experiments demonstrate that DCPI can be seamlessly integrated into state-of-the-art DC methods. Particularly, for coreset selection methods, applying DCPI to Herding on CIFAR10 (with a fraction of 0.4\%) improves performance by 24.3\%, while applying it to K-center in cross-architecture evaluations leads to an improvement of up to 23.4\%. For dataset distillation methods, integrating DCPI with the DC method on CIFAR100 (with a fraction of 0.2\%) yields a 2.1\% improvement, and further cross-architecture evaluations of DC show gains of up to 18.3\%.
\end{enumerate}

\section{Related Work}
\label{sec:relatedwork}
Following prior work on dataset condensation (DC)~\cite{herding,DC,LCMat}, we consider a multi-classification problem. Let $\mathcal{X} \in \mathbb{R}^d$ be the input space and $\mathcal{Y}$ the label set. The dataset $\mathcal{D}_{\mathcal{T}} = \{(x_i, y_i)\}_{i=1}^{n} \subseteq \mathcal{X} \times \mathcal{Y}$ has $n$ samples, with $x_i \in \mathcal{X}$ as input vectors and $y_i \in \mathcal{Y}$ as labels. DR aims to create a smaller dataset $\mathcal{D}_{\mathcal{S}}$ of size $m \ll n$. DR methods include \textit{coreset selection}, using a subset of the original data, and \textit{dataset distillation}, using optimized synthetic data not in the original set.

\textbf{Coreset Selection Techniques.} Coreset selection identifies a representative subset $\mathcal{D}_{\mathcal{S}}$ from dataset $\mathcal{D}_{\mathcal{T}}$, optimizing a criterion for informativeness matching $\mathcal{D}_{\mathcal{T}}$. Informativeness is measured by metrics like gradients~\cite{Data-Diet,Craig,gradmatchcoreset}, loss values~\cite{forgetting}, predictive uncertainties~\cite{Uncertainty}, decision boundary proximity~\cite{DeepFool,CAL}, or model sharpness~\cite{LCMat}.

\textbf{Dataset Distillation Methods.} Dataset distillation synthesizes a distilled dataset $\mathcal{D}_{\mathcal{S}}$ from $\mathcal{D}_{\mathcal{T}}$ via bi-level optimization to match $\mathcal{D}_{\mathcal{T}}$'s performance. A distance metric $\mathbf{D}$ measures statistical divergence, guiding updates: $\mathcal{D}_{\mathcal{S}} \leftarrow \mathcal{D}_{\mathcal{S}} - \eta \cdot \nabla_{\mathcal{D}_{\mathcal{S}}} \mathbf{D}(\mathcal{D}_{\mathcal{S}}, \mathcal{D}_{\mathcal{T}})$. Metric $\mathbf{D}$ may include gradients~\cite{DC,DCC,DSA}, features~\cite{DM,DataDAM}, training trajectories~\cite{MTT,FTD,TESLA,DATM}, or kernel information~\cite{KIP,KIP_inf,FRePo}.

In addition to direct performance matching, certain methodologies endeavor to re-parameterize input data to enhance compression efficiency. Techniques employed in this context include exploiting data regularity, as discussed in various studies~\cite{IDC,IDM,FYI}, factorizing images to capture intrinsic structures~\cite{Haba,Linba}, and employing sparse coding to represent data effectively~\cite{speed}. 

\section{Methodology}
\label{sec:Methodology}




\begin{figure}[tb!]
    \centering
    \includegraphics[width=0.99\linewidth]{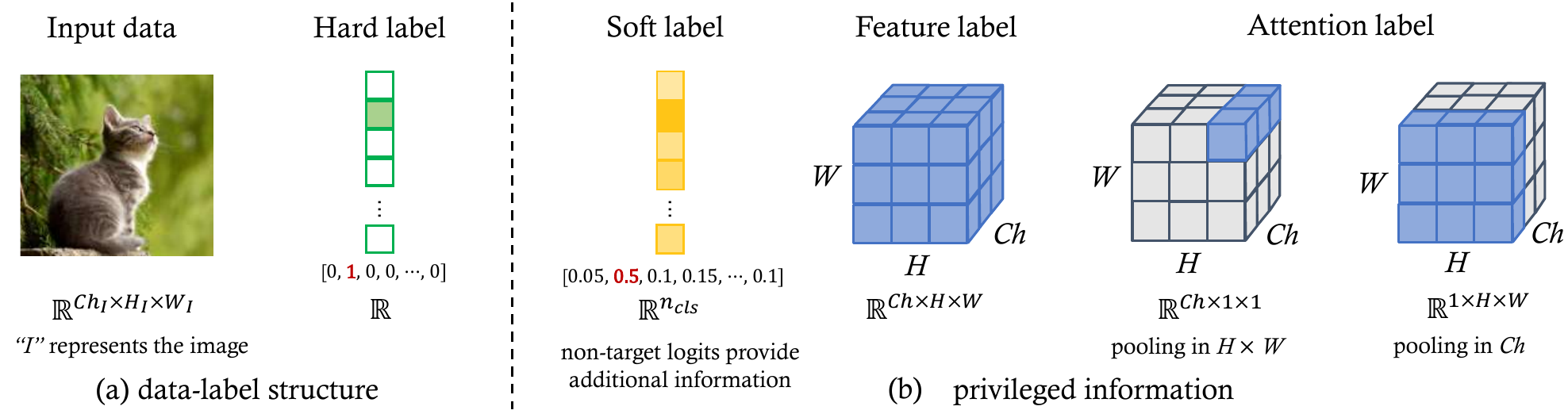}
    \vspace{-5pt}
    \caption{Comparison between (a) the traditional ``data-label'' structure and (b) Different forms of privileged information. Non-target classes of soft labels provide additional information, can be considered a form of privileged information. Feature labels encapsulate high-dimensional information. Attention labels are obtained by applying average pooling to feature labels.}
    \label{fig:PI_format}
    \vspace{-12pt}
\end{figure}

\subsection{Determining Privileged Information}
\label{sec:Id_PI}
Although prior research on dataset condensation has shown impressive results, it mainly focuses on generating reduced datasets in conventional data-label formats, as depicted in Figure~\ref{fig:PI_format}(a).
However, more informative data, such as privileged information, can be utilized to enhance both the utility and representational performance of reduced datasets, as illustrated in Figure~\ref{fig:PI_format}(b). Below, we briefly explore several forms of privileged information that can be incorporated to achieve this goal.

\noindent \textbf{Soft labels.} We first claim that soft labels are a form of privileged information, as they offer richer insight into how an expert model interprets predictions, providing soft probabilities rather than a single hard label. Specifically, the non-target logits can be viewed as additional information for supervision. Several works have previously explored the effectiveness of soft labels in dataset condensation~\cite{DATM,TESLA,LD,nodistill}. However, while soft labels enhance the available information, they are limited to low-dimensional discriminative representations and fail to capture more complex, high-dimensional statistics. Moreover, they do not alter the data-label structure of the reduced dataset representation. 

\noindent \textbf{Feature labels.} Beyond soft labels, we propose feature labels as a more effective form of privileged information. These labels, derived from unified intermediate representations across well-trained models, encapsulate rich, high-dimensional latent statistics. By providing additional supervision, feature labels enhance model training on datasets with privileged information. Unlike approaches that focus primarily on soft labels, assigning a unified feature label to each input enriches the supervision signal for downstream tasks, effectively addressing the limitations of prior methods.

\noindent \textbf{Attention labels.} Alongside feature labels, we propose attention labels as an alternative form of privileged information that provides a more memory-efficient representation. Attention labels can be derived from either spatial or channel attention of feature labels~\cite{woo2018cbam}. For example, given a feature label of size $Ch \times H \times W$, spatial attention reduces the $Ch$ dimensions through pooling operations (\emph{e.g.}, average pooling, max pooling), resulting in an attention label of size $1 \times H \times W$. Similarly, channel attention applies pooling along the $H\times W$ dimension to produce a reduced representation, \emph{i.e.}, $Ch\times 1 \times 1$. Both feature and attention labels are valuable, with attention labels offering a more efficient representation but with a possible trade-off in the richness of information. 

In this work, we primarily focus on generating additional \textit{feature labels}\footnote{We consider attention labels as a specific form of feature labels. Therefore, for simplicity, we use the term “feature labels” as a unified description for both feature labels and attention labels.} for the reduced dataset, as these forms of privileged information provide more complementary and useful insights for model training. However, privileged information can take various forms beyond attention and feature labels. Depending on the task and the model architecture, other types of information, such as learned embeddings, domain-specific signals, or task-related metadata, could be equally beneficial in enhancing the informativeness and performance of reduced datasets. Incorporating diverse privileged information lets us tailor datasets to specific needs and maximize reduced data potential.
\vspace{-10pt}
\subsection{Synthesizing Privileged Information}

\label{sec:Syn_PI}
In this section, we discuss the process of generating privileged information for a given reduced dataset $\mathcal{D}_{\mathcal{S}} = \{(\tilde{x}_i, \tilde{y}_i)\}_{i=1}^m$, with the goal of obtaining a more informative dataset $\mathcal{D}_{\mathcal{S}}^\star = \{(\tilde{x}_i, \tilde{y}_i, f_i^\star)\}_{i=1}^m$. Here, $\mathcal{D}_{\mathcal{S}}$ is the reduced dataset of a larger dataset $\mathcal{D}_{\mathcal{T}} = \{(x_i, y_i)\}_{i=1}^n$, where $m \ll n$. Our primary focus is on incorporating feature labels as the form of privileged information. While various methods can be employed to synthesize privileged information, they can generally be categorized into two strategies: \textit{direct assignment} and \textit{learning-based methods.}

\noindent \textbf{Direct Assignment of Feature Labels.}
A straightforward approach to obtaining feature labels is by using a pre-trained model, \emph{e.g.}, $g_{\mathcal{T}}$, to extract intermediate features for each input data $\tilde{x}_i \in \mathcal{D}_{\mathcal{S}}$. Specifically, this is formalized as $f_i^\star = g_{\mathcal{T}}(\tilde{x}_i)$, resulting in an extended dataset represented as $\mathcal{D}_{\mathcal{S}}^\star = {(\tilde{x}_i, \tilde{y}_i, f_i^\star)}$. This method is computationally efficient but relies heavily on the generalization ability of the pre-trained model $g_{\mathcal{T}}$. While the feature labels, which capture the implicit biases of $g_{\mathcal{T}}$, may enhance the quality of the reduced dataset, they could also introduce potential drawbacks. In fact, directly assigned feature labels are often overly discriminative, reducing diversity. However, we find that suitable feature labels should strike a balance between these two properties.

\noindent \textbf{Learning Feature Labels.}
A more robust approach to obtaining feature labels is through learning-based methods. Many dataset distillation techniques can be adapted for learning feature labels. For instance, we can employ the Dataset Condensation (DC)~\cite{DC} method as an illustrative example to guide the process of learning synthetic feature labels. Suppose we have a learned synthetic dataset $\mathcal{D}_{\mathcal{S}}$. In the typical DC method, we initialize a random model $g$ with parameters $\theta = \theta_0$ and train it for $T$ epochs on both $\mathcal{D}_{\mathcal{T}}$ and $\mathcal{D}_{\mathcal{S}}$ separately. The synthetic dataset $\mathcal{D}_{\mathcal{S}}$ is updated by matching the category gradients between the two datasets, which can be expressed as follows\footnote{We ignore the category symbol for simplicity.}:
\vspace{-5pt}
\begin{equation}
\label{eq:DC}
\begin{aligned}
    \mathcal{D}_{\mathcal{S}} = \mathop{\arg\min}_{\mathcal{D}_{\mathcal{S}}} & \underset{\theta_0 \sim P_\theta}{\mathbb{E}}
\left[ \sum_{t=0}^T \mathbf{D}\left(\nabla_\theta L(\mathcal{D}_{\mathcal{S}};\theta_t), \nabla_\theta L(\mathcal{D}_{\mathcal{T}};\theta_t)\right)\right] \\
\text{where} \quad  & L(\mathcal{D}_{\mathcal{T}};\theta_t)  = \underset{(x_i,y_i)\in \mathcal{D}_{\mathcal{T}}}{\mathbb{E}} \ell_{ce}\left[\left(y_i, \sigma(g(x_i;\theta_t))\right)\right], \\
\text{and} \quad & L(\mathcal{D}_{\mathcal{S}};\theta_t)\triangleq \mathcal{L}_{cls} =\underset{ (\tilde{x}_i, \tilde{y}_i) \in \mathcal{D}_{\mathcal{S}}}{\mathbb{E}} \left[\ell_{ce}\left(\tilde{y}_i,\sigma(g(\tilde{x}_i;\theta_t))\right)\right],
\end{aligned}
\end{equation}

where $\ell_{ce}(\cdot, \cdot)$ denotes the cross-entropy (CE) loss, and $\sigma(\cdot)$ represents the softmax function. In contrast, we aim to match the performance between $\mathcal{D}_{\mathcal{T}}$ and $\mathcal{D}_{\mathcal{S}}^\star$, where privileged information is synthesized to capture additional informativeness. Let $\ell_{mse}$ represent the mean square error (MSE) loss, and let $\psi(\cdot)$ denote the intermediate output of model $g$, \emph{i.e.}, $g = \psi \circ \kappa$, where $\kappa(\cdot)$ is the classifier component of $g$. Therefore, our objective becomes:
\begin{equation}
    \mathcal{D}_{\mathcal{S}}^\star  =  \mathop{\arg\min}_{\mathcal{D}_{\mathcal{S}}^\star}   \underset{\theta_0 \sim P_\theta}{\mathbb{E}}  \left[  \sum_{t=0}^T \mathbf{D}  (  \nabla_\theta L_c(\mathcal{D}_{\mathcal{S}}^\star;\theta_t), \nabla_\theta L_c(\mathcal{D}_{\mathcal{T}};\theta_t))\right],
\end{equation}
\begin{equation}
\begin{aligned}
\label{eq:DC_PI}
     \text{where}  \quad &  L(\mathcal{D}_{\mathcal{S}}^\star;\theta_t)  \triangleq \mathcal{L}_{cls} + \lambda_{reg} \cdot \mathcal{L}_{reg}, \\
     & \mathcal{L}_{cls}  = \underset{(\tilde{x}_i,\tilde{y}_i)\in \mathcal{D}_{\mathcal{S}}^\star}{\mathbb{E}} [\ell_{ce}\left(\tilde{y}_i, \sigma(g(\tilde{x}_i;\theta_t))\right)],  \\ 
     \text{and} \quad & \mathcal{L}_{reg}  = \underset{(\tilde{x}_i,\tilde{y}_i,f_i^\star)\in \mathcal{D}_{\mathcal{S}}^\star}{\mathbb{E}} [\ell_{mse}\left(f_i^\star,\psi(\tilde{x}_i;\theta_t)\right)],
\end{aligned}
\end{equation}
where $\lambda_{reg}$ is a hyper-parameter to determine the scale of using privileged information. In addition to DC, other dataset distillation methods can also be used synthesize feature labels $f_i^\star$. We provide further results on coreset selection methods like Herding~\cite{herding}, K-center~\cite{k_center}, Forgetting~\cite{forgetting}, and dataset distillation methods like DC~\cite{DC}, MTT~\cite{MTT}, and DATM~\cite{DATM}.

In addition, we introduce additional supervision to enhance the discriminative power of these feature labels while preserving their diversity.

\begin{figure}[tb!]
    \centering
\includegraphics[width=0.99\linewidth]{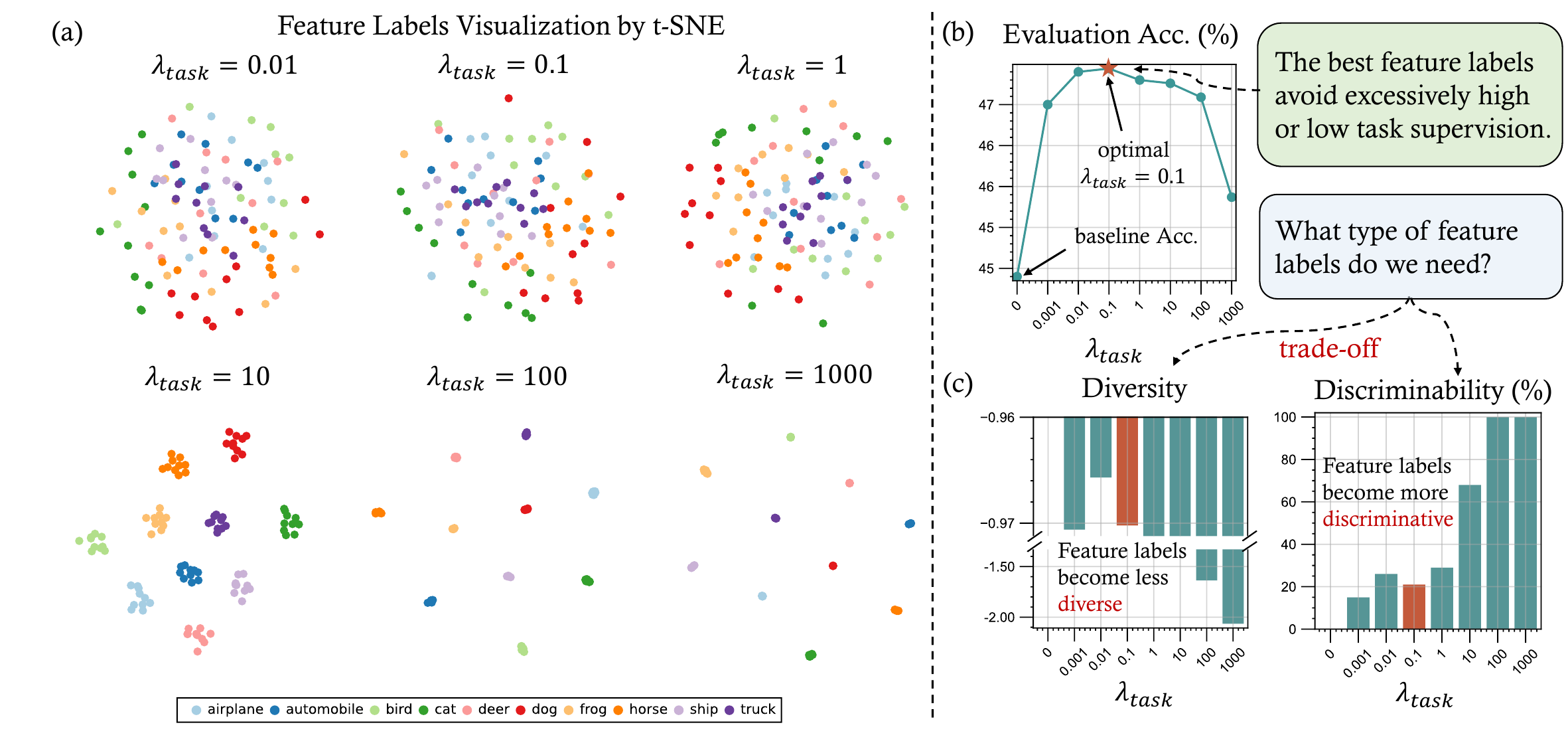}
    \caption{Feature labels learned under varying levels of task supervision. (a) t-SNE visualization of feature labels learned with different task supervision coefficients $\lambda_{task}$. (b) The most effective feature labels are produced with a moderate level of task supervision, avoiding excessively high or low supervision. (c) Increasing task supervision makes the feature labels more discriminative but less diverse. Diversity is measured by the negative mutual information between the feature labels and the ground truth labels, while discriminability is measured by the classification accuracy of a linear classifier trained on the feature labels.}
    \label{fig:ce_role}
    \vspace{-19pt}
\end{figure}

\noindent \textit{$\bullet$ Task-oriented synthesization.}
To improve the discriminative power of the feature labels, we adopt a task-oriented approach by feeding the synthesized feature labels into the classifier of the model used to extract gradients during bi-level optimization. We achieve this by performing additional CE loss between the feature label $f_i^\star$'s prediction and the ground-truth label $\tilde{y}_i$. Specifically, we have
\begin{equation}
\label{eq:loss_task}
    \mathcal{L}_{task}  = \underset{(\tilde{f}_i^\star,\tilde{y}_i)\in \mathcal{D}_{\mathcal{S}}^\star}{\mathbb{E}} [\ell_{ce}\left(\tilde{y}_i,\sigma(\kappa(f_i^\star;\theta_t))\right)].
\end{equation}
This allows the feature labels to contribute directly to the final prediction, ensuring they become better aligned with the task at hand. The scale of task supervision is controlled by the hyper-parameter $\lambda_{task}$. Our observations indicate that the \textbf{preferred feature labels should strike a balance between discriminability and diversity, \emph{i.e.}, they should neither be overly discriminative nor completely lack discriminative power}. As shown in Figure~\ref{fig:ce_role}(a)(c), increasing $\lambda_{task}$ tends to cluster the feature labels, reducing their diversity while increasing their discriminability. We find that the optimal feature labels are not achieved with the highest task supervision. Instead, a moderate level of task supervision, as shown in Figure~\ref{fig:ce_role}(b), strikes the right balance between diversity and discriminability. This also suggests a possible explanation for the drawbacks of directly assigning feature labels from a well-trained model, as these labels tend to be overly discriminative.

\noindent \textit{$\bullet$ Versatility synthesization.}
Beyond task-specific supervision, we aim to preserve the generative versatility of the feature labels, \emph{i.e.}, a set of feature labels $f_i^\star \in F_i$, where $F_i$ represents the possible feature label set for input $x_i$. This approach involves synthesizing multiple feature labels for a single data-label pair. Appropriate versatility enhancement ensures that the synthesized feature labels remain informative across different tasks and applications, providing a richer and more comprehensive representation of the data. When using multiple feature labels, we primarily employ two strategies: randomly selecting one or using the average feature label from $F_i$. Further discussions demonstrating the benefits of this versatility are presented in Figure~\ref{fig:ablation}(a) and Table~\ref{tab:cross_all} in Appendix~\ref{app:supervision_n_features}. 

We define the overall loss function for a model trained on the reduced dataset $\mathcal{D}_{\mathcal{S}}^\star$ as follows:
\begin{equation}
\label{eq:loss}
    L(\mathcal{\mathcal{D}_{\mathcal{S}}^\star};\theta_t) = \mathcal{L}_{cls} + \mathbb{E}_{f_i^\star\in F_i}\left[\lambda_{reg}\cdot \mathcal{L}_{reg} + \lambda_{task}\cdot \mathcal{L}_{task}\right],
\end{equation}
where $F_i$ denotes the feature label set, containing multiple $f_i^\star$ for a single data-label pair $(\tilde{x}_i, \tilde{y}_i)$. The pseudocode for learning privileged information is provided in Algorithm~\ref{alg:process} in Appendix~\ref{app:pseudo_code}.

\subsection{Learning Using Privileged Information}
\label{sec:LUPI}
As discussed above, we have discussed how to synthesize appropriate privileged information $f_i^\star$ for a given reduced dataset $\mathcal{D}_{\mathcal{S}}$ and extend it into $\mathcal{D}_{\mathcal{S}}^\star$. We now focus on leveraging the new reduced dataset $\mathcal{D}_{\mathcal{S}}^\star$ to enhance a model’s performance on unseen test data. Following the \textit{learning using privileged information} (LUPI) framework~\cite{privileged_info,unify_privileged_info}, we incorporate the additional privileged information $f_i^\star$ during training to build a classifier that outperforms those trained solely on the regular reduced dataset $\mathcal{D}_{\mathcal{S}}$.

Given an arbitrary model $h$ with parameters $\theta\in \Theta$, we now formally discuss how to train $h$ with synthesized feature labels. The same loss function $L(\mathcal{D}_{\mathcal{S}}^\star)$, as shown in Eq.~(\ref{eq:loss}), is applied to train $h$, with hyper-parameters (\emph{e.g.}, $\lambda_{reg},\lambda_{task}$) kept consistent for fairness. 
\begin{equation}
    \theta^\star = \mathop{\arg\min}_{\theta\in \Theta} L(\mathcal{D}_{\mathcal{S}}^\star;\theta)
\end{equation}
Besides feature labels, we can store attention labels, which can be generated by performing average pooling on the spatial or channel dimensions of learned feature labels. Attention labels contain more condensed information, which can further reduce storage cost. During LUPI, we apply the same pooling strategy for intermediate features of the given model to calculate the MSE loss between the intermediate features and given feature labels. 

\section{Experiments}
\label{sec:Experiments}

\begin{table*}[tb!]
\centering
\vspace{-5pt}
\caption{Application of DCPI to representative pruning methods on CIFAR-10/100. We initialized the dataset using the baseline methods and utilized DC for synthesizing feature labels, with the same memory cost for fair compairason.}
\vspace{-10pt}
\label{tab:corset_main}
\resizebox{0.99\textwidth}{!}{
\begin{tabular}{c|*{5}{c}|c*{4}{c}}
\toprule
 Dataset & \multicolumn{5}{c|}{CIFAR-10} & \multicolumn{5}{c}{CIFAR-100} \\ 
Fraction (\%) & 0.02 & 0.1 & 0.2 & 0.4 & 1 & 0.2 & 1 & 2 & 4 & 10 \\ \midrule
Random & 13.5{\scriptsize$\pm$0.4 }&20.0{\scriptsize$\pm$0.5 } &27.1{\scriptsize$\pm$0.6 }  & 35.8{\scriptsize$\pm$0.6 } &  43.0{\scriptsize$\pm$0.5 }
&4.3{\scriptsize$\pm$0.2 }  & 9.5{\scriptsize$\pm$0.2 } &14.5{\scriptsize$\pm$0.2 } &19.5{\scriptsize$\pm$0.4 }  & 29.5{\scriptsize$\pm$0.3 }\\ \midrule
L-Conf &10.7{\scriptsize$\pm$0.4 } & 10.5{\scriptsize$\pm$0.4 }&10.8{\scriptsize$\pm$0.4 }  &17.9{\scriptsize$\pm$0.4 }  &  23.1{\scriptsize$\pm$0.6 }
&2.1{\scriptsize$\pm$0.1 }  &3.6{\scriptsize$\pm$0.1 }  &6.6{\scriptsize$\pm$0.2 } & 9.0{\scriptsize$\pm$0.2 } &16.4{\scriptsize$\pm$0.3 } \\ 
Entropy &12.2{\scriptsize$\pm$0.6 } &14.1{\scriptsize$\pm$0.5 } &14.8{\scriptsize$\pm$0.5 }  &19.6{\scriptsize$\pm$0.4 }  &  23.8{\scriptsize$\pm$0.7 }
& 1.7{\scriptsize$\pm$0.1 } &3.7{\scriptsize$\pm$0.2 }  &6.7{\scriptsize$\pm$0.2 }& 9.0{\scriptsize$\pm$0.3 } & 17.1{\scriptsize$\pm$0.3 }\\ 
Margin &8.9{\scriptsize$\pm$0.4 } &15.8{\scriptsize$\pm$0.7 } &20.3{\scriptsize$\pm$0.4 }  &24.8{\scriptsize$\pm$0.5 }  &  31.3{\scriptsize$\pm$0.5 }
&3.0{\scriptsize$\pm$0.2 }  & 6.2{\scriptsize$\pm$0.2 } &9.0{\scriptsize$\pm$0.2 } & 12.7{\scriptsize$\pm$0.3 } & 20.7{\scriptsize$\pm$0.3 }\\ 
Glister &11.5{\scriptsize$\pm$0.3 } &16.9{\scriptsize$\pm$0.5 } &23.0{\scriptsize$\pm$0.4 }  &28.4{\scriptsize$\pm$0.5 }  & 30.0{\scriptsize$\pm$0.5 }
&2.9{\scriptsize$\pm$0.5 }  & 7.1{\scriptsize$\pm$0.4 } &10.4{\scriptsize$\pm$0.6 } &13.3{\scriptsize$\pm$0.7 }  &27.2{\scriptsize$\pm$0.7 }\\ 
Graig &18.1{\scriptsize$\pm$0.3 } &19.5{\scriptsize$\pm$0.4 } &19.0{\scriptsize$\pm$0.5 }  &27.8{\scriptsize$\pm$0.3 }  &30.2{\scriptsize$\pm$0.4 }  
&4.3{\scriptsize$\pm$0.5 }  &9.0{\scriptsize$\pm$0.4 }  &13.6{\scriptsize$\pm$0.7 } &14.6{\scriptsize$\pm$0.5 }  &20.1{\scriptsize$\pm$0.6 } \\ \midrule
Herding & 15.3{\scriptsize$\pm$0.5 } & 23.5{\scriptsize$\pm$0.3 } & 25.1{\scriptsize$\pm$0.5 } & 26.7{\scriptsize$\pm$0.4 } & 34.9{\scriptsize$\pm$0.6 } & 4.0{\scriptsize$\pm$0.1 } & 6.2{\scriptsize$\pm$0.2 } & 8.1{\scriptsize$\pm$0.4 } & 13.1{\scriptsize$\pm$0.6 } & 18.4{\scriptsize$\pm$0.6 } \\ 
\rowcolor{cyan!10}+DCPI & \textbf{27.9{\scriptsize$\pm$0.6 }} & \textbf{37.3{\scriptsize$\pm$0.6 }} & \textbf{45.8{\scriptsize$\pm$0.7 }} & \textbf{51.0{\scriptsize$\pm$0.4 }} & \textbf{54.0{\scriptsize$\pm$0.6 }} & \textbf{14.0{\scriptsize$\pm$0.3 }} & \textbf{20.4{\scriptsize$\pm$0.5 }} & \textbf{25.5{\scriptsize$\pm$0.4 }} & \textbf{28.9{\scriptsize$\pm$0.3 }} & \textbf{31.4{\scriptsize$\pm$0.6 }} \\ 
\midrule
k-Center & 16.4{\scriptsize$\pm$0.6 } & 22.4{\scriptsize$\pm$0.6 } & 23.1{\scriptsize$\pm$0.5 } & 30.4{\scriptsize$\pm$0.4 } & 36.7{\scriptsize$\pm$0.5 } & 4.8{\scriptsize$\pm$0.2 } & 6.7{\scriptsize$\pm$0.4 } & 10.0{\scriptsize$\pm$0.5 } & 15.9{\scriptsize$\pm$1.1 } & 21.8{\scriptsize$\pm$1.0 } \\ 
\rowcolor{cyan!10}+DCPI & \textbf{29.7{\scriptsize$\pm$0.6 }} & \textbf{40.0{\scriptsize$\pm$0.6 }} & \textbf{46.2{\scriptsize$\pm$0.6 }} & \textbf{50.8{\scriptsize$\pm$0.6 }} & \textbf{54.3{\scriptsize$\pm$0.6 }} & \textbf{13.5{\scriptsize$\pm$0.3 }} & \textbf{20.0{\scriptsize$\pm$0.5 }} & \textbf{25.9{\scriptsize$\pm$0.3 }} & \textbf{29.1{\scriptsize$\pm$0.5 }} & \textbf{32.0{\scriptsize$\pm$0.5 }} \\ 
\midrule
Forgetting & 15.3{\scriptsize$\pm$0.6 } & 19.1{\scriptsize$\pm$0.7 } & 23.9{\scriptsize$\pm$0.7 } & 26.9{\scriptsize$\pm$0.7 } & 39.5{\scriptsize$\pm$0.5 } & 4.1{\scriptsize$\pm$0.1 } & 7.8{\scriptsize$\pm$0.3 } & 10.4{\scriptsize$\pm$0.5 } & 14.1{\scriptsize$\pm$0.6 } & 22.3{\scriptsize$\pm$0.4 } \\ 
\rowcolor{cyan!10}+DCPI & \textbf{30.0{\scriptsize$\pm$0.6 }} & \textbf{39.7{\scriptsize$\pm$0.7 }} & \textbf{46.6{\scriptsize$\pm$0.6 }} & \textbf{51.3{\scriptsize$\pm$0.5 }} & \textbf{54.5{\scriptsize$\pm$0.5 }} & \textbf{14.0{\scriptsize$\pm$0.4 }} & \textbf{20.1{\scriptsize$\pm$0.6 }} & \textbf{25.8{\scriptsize$\pm$0.3 }} & \textbf{29.3{\scriptsize$\pm$0.4 }} & \textbf{32.2{\scriptsize$\pm$0.5 }} \\ 
\midrule
 Full Dataset & \multicolumn{5}{c|}{84.8{\scriptsize$\pm$0.1 }} & \multicolumn{5}{c}{56.2{\scriptsize$\pm$0.3 }}  \\ 
\bottomrule
\end{tabular}
}
\end{table*}

\begin{table*}[tb!]
\centering
\vspace{-5pt}
\caption{Results of DCPI on distillation methods across CIFAR-10/100, and Tiny ImageNet. We initialized reduced datasets with corresponding baseline methods, and synthesized feature labels for these datasets with the same baseline methods.}
\vspace{-10pt}
\label{tab:dd_main_exp}
{\resizebox{0.99\textwidth}{!}{
\begin{tabular}{c|*{3}{c}|c*{2}{c}|c*{2}{c}}
\toprule
 Dataset & \multicolumn{3}{c|}{CIFAR-10} & \multicolumn{3}{c|}{CIFAR-100} & \multicolumn{2}{c}{Tiny ImageNet} \\ 
Fraction (\%) & 0.02 & 0.2 & 1 & 0.2 & 2 & 10 & 0.2 & 2  \\ \midrule
Random & 15.4{\scriptsize$\pm$0.3 } & 31.0{\scriptsize$\pm$0.5 } & 50.6{\scriptsize$\pm$0.3 } & 4.2{\scriptsize$\pm$0.3 } & 14.6{\scriptsize$\pm$0.5 } & 33.4{\scriptsize$\pm$0.4 } & 1.4{\scriptsize$\pm$0.1 } & 5.0{\scriptsize$\pm$0.2 } \\ \midrule
KIP & 49.9{\scriptsize$\pm$0.2 } & 62.7{\scriptsize$\pm$0.3 } & 68.6{\scriptsize$\pm$0.3 } & 15.7{\scriptsize$\pm$0.2 } & 28.3{\scriptsize$\pm$0.1 } & - & - & - \\ 
DM & 26.0{\scriptsize$\pm$0.8 } & 48.9{\scriptsize$\pm$0.6 } & 63.0{\scriptsize$\pm$0.4 } & 11.4{\scriptsize$\pm$0.3 } & 29.7{\scriptsize$\pm$0.3 } & 43.6{\scriptsize$\pm$0.4 } & 3.9{\scriptsize$\pm$0.2 } & 12.9{\scriptsize$\pm$0.4 } \\ 
DSA & 28.8{\scriptsize$\pm$0.7 } & 52.1{\scriptsize$\pm$0.5 } & 60.6{\scriptsize$\pm$0.5 } & 13.9{\scriptsize$\pm$0.3 } & 32.3{\scriptsize$\pm$0.3 } & 42.8{\scriptsize$\pm$0.4 } & - & - \\ 
DCC & 32.9{\scriptsize$\pm$0.8 } & 49.4{\scriptsize$\pm$0.5 } & 61.6{\scriptsize$\pm$0.4 } & 13.3{\scriptsize$\pm$0.3 } & 30.6{\scriptsize$\pm$0.4 } & 40.0{\scriptsize$\pm$0.3 } & - & - \\ 
DSAC & 34.0{\scriptsize$\pm$0.7 } & 54.5{\scriptsize$\pm$0.5 } & 64.2{\scriptsize$\pm$0.4 } & 14.6{\scriptsize$\pm$0.3 } & 33.5{\scriptsize$\pm$0.3 } & 39.3{\scriptsize$\pm$0.4 } & - & - \\ 
CAFE & 30.3{\scriptsize$\pm$1.1 } & 46.3{\scriptsize$\pm$0.6 } & 55.5{\scriptsize$\pm$0.6 } & 12.9{\scriptsize$\pm$0.3 } & 27.8{\scriptsize$\pm$0.3 } & 37.9{\scriptsize$\pm$0.3 } & - & - \\ 
IDM & 45.6{\scriptsize$\pm$0.7 } & 58.6{\scriptsize$\pm$0.1 } & 67.5{\scriptsize$\pm$0.1 } & 20.1{\scriptsize$\pm$0.3 } & 45.1{\scriptsize$\pm$0.1 } & 50.0{\scriptsize$\pm$0.2 } & - & - \\ 
\midrule
DC & 28.3{\scriptsize$\pm$0.5 } & 44.9{\scriptsize$\pm$0.5 } & 53.9{\scriptsize$\pm$0.5 } & 12.8{\scriptsize$\pm$0.3 } & 25.2{\scriptsize$\pm$0.3 } & 29.8{\scriptsize$\pm$0.3 } & - & - \\ 
 \rowcolor{cyan!10}+DCPI & 31.5{\scriptsize$\pm$0.9 } & 47.3{\scriptsize$\pm$0.9 } & 55.3{\scriptsize$\pm$0.5 } & 14.9{\scriptsize$\pm$0.4 } & 25.6{\scriptsize$\pm$0.5 }& 31.6{\scriptsize$\pm$0.5 } & - & -  \\
\midrule
MTT & 46.2{\scriptsize$\pm$0.8 } & 65.4{\scriptsize$\pm$0.7 } & 71.6{\scriptsize$\pm$0.2 } & 24.3{\scriptsize$\pm$0.3 } & 39.7{\scriptsize$\pm$0.4 } & 47.7{\scriptsize$\pm$0.2 } & 8.8{\scriptsize$\pm$0.3 } & 23.2{\scriptsize$\pm$0.2 } \\ 
\rowcolor{cyan!10}+DCPI & \textbf{47.4{\scriptsize$\pm$0.5 }} & \textbf{65.9{\scriptsize$\pm$0.6 }} & \textbf{72.7{\scriptsize$\pm$0.2 }} & 25.6{\scriptsize$\pm$0.4 }& 41.0{\scriptsize$\pm$0.3 } & 48.4{\scriptsize$\pm$0.3 } & 11.2{\scriptsize$\pm$0.1 } & 24.9{\scriptsize$\pm$0.2 } \\
\midrule


RDED & 23.5{\scriptsize$\pm$0.3 } & 50.2{\scriptsize$\pm$0.3 } & 68.4{\scriptsize$\pm$0.1 } & 19.6{\scriptsize$\pm$0.3 } & 48.1{\scriptsize$\pm$0.3 }& 57.0{\scriptsize$\pm$0.1 }  & 12.0{\scriptsize$\pm$0.1 } & 39.6{\scriptsize$\pm$0.1 }\\ 
\rowcolor{cyan!10}+DCPI & 28.2{\scriptsize$\pm$0.4 } & 50.6{\scriptsize$\pm$0.3 } & 
68.5{\scriptsize$\pm$0.1 } & \textbf{32.5{\scriptsize$\pm$0.4 }} & \textbf{48.3{\scriptsize$\pm$0.4 }} & \textbf{57.1{\scriptsize$\pm$0.1 }} & \textbf{15.2{\scriptsize$\pm$0.3 }} & \textbf{40.8{\scriptsize$\pm$0.2 }}    \\  
\midrule
 Full Dataset & \multicolumn{3}{c|}{84.8{\scriptsize$\pm$0.1 }} & \multicolumn{3}{c|}{56.2{\scriptsize$\pm$0.3 }} & \multicolumn{2}{c}{37.6{\scriptsize$\pm$0.4 }} \\ 
\bottomrule
\end{tabular}}
}
\vspace{-10pt}
\end{table*}

\subsection{Experimental Setup}
\label{sec:exp_implementation}
In this section, we investigate the effectiveness of our proposed method, DCPI, through a series of experiments on diverse datasets and tasks. We begin by evaluating the efficacy of DCPI when applied to coreset selection and dataset distillation tasks. Specifically, we followed prior works to conduct experiments on  CIFAR-10/100~\cite{krizhevsky2009learning} for coreset selection methods, where ResNet-18~\cite{resnet} is utilized for extracting importance score. For the dataset distillation methods, we conducted experiments on CIFAR-10/100, Tiny ImageNet~\cite{le2015tiny}, and subsets of ImageNet~\cite{ILSVRC15}.

For coreset selection, we benchmarked our method against several representative baselines, including Random, L-conf, Entropy, Margin~\cite{Uncertainty}, Glister~\cite{glister}, Graig~\cite{Craig}, Herding~\cite{herding}, k-Center~\cite{k_center}, and Forgetting~\cite{forgetting}. More detailed settings are provided in Appendix~\ref{app:prune}. 

For dataset distillation, we evaluated advanced methods, including KIP~\cite{KIP},  DM~\cite{DM}, DSA~\cite{DSA}, DCC~\cite{lee2022dataset}, DSAC~\cite{DCC}, CAFE~\cite{CAFE}, IDM~\cite{IDM},  DC~\cite{DC}, and MTT~\cite{MTT}. In line with prior studies, we used networks with instance normalization as the default setting. Unless otherwise specified, distillation was performed with a depth-3 ConvNet for CIFAR-10/100, a depth-4 ConvNet for Tiny ImageNet, and a depth-5 ConvNet for ImageNet subsets. Table~\ref{tab:dd_setting_cifar_tiny} and Table~\ref{tab:dd_setting_imagenet} in Appendix~\ref{app:distillation} provided detailed information regarding hyperparameter settings.

It is worth noting that for both pruning and distillation methods, we initialized all data-label pairs using the baseline method and employed a weakly-trained model (trained for one epoch) to extract feature labels, which were then used to synthesize privileged information. By default, we utilized DC to synthesize one feature label per data-label pair, aligning it with features extracted from the final layer of a ConvNet during bi-level optimization.  
\vspace{-8pt}

\subsection{Main Results}
\label{sec:Main_results}
\noindent \textbf{Coreset selection.} In our experiments, DCPI utilizes the reduced dataset  initialized with the Herding, k-Center, and Forgetting methods to assess its performance across diverse fraction on CIFAR-10/100. \textbf{To ensure a fair comparison, our experiments in this work are conducted under the same memory constraints as the baseline methods.} As shown in Table~\ref{tab:corset_main}, by incorporating privileged information, these methods consistently outperformed the baseline across a range of fraction settings on CIFAR-10/100. Particularly, on the CIFAR-10 (fraction = 0.4\%) , DCPI achieved a performance increase of 24.4\% on the Forgetting method and 24.3\% on the Herding method. We find that for datasets without optimized instances (\emph{e.g.}, selected coresets), the performance gain is much higher than that with optimized samples.


\noindent \textbf{Dataset distillation.} We employed DCPI in several methods, where privileged information is obtained with DC and MTT and RDED. Table~\ref{tab:dd_main_exp} summarizes the classification performances of ConvNets trained with different distillation methods. Specifically, applying DCPI to DC on CIFAR-100 (fraction = 0.2\%) resulted in a 2.1\% improvement. For MTT, DCPI delivered a 2.4\% gain on Tiny ImageNet (fraction = 0.2\%). Additionally, we evaluated its effectiveness on ImageNet subsets, as shown in Table~\ref{tab:mtt_subsets}, where DCPI applied to MTT led to a 3.4\% improvement on ImageMeow with fraction = 1\%, demonstrating strong performance even on larger datasets. Notably, learning both feature labels with DCPI outperforms simply extracting features alone. Further results for DATM are provided in Appendix~\ref{app:further_results}. The application of DCPI to RDED is detailed in Tables~\ref{tab:dd_main_exp} and~\ref{tab:rded}. Specifically, DCPI achieved a 12.9\% improvement on CIFAR-100 (fraction = 0.2\%). Furthermore, its effectiveness extends to large-scale datasets like ImageNet-1k. For instance, when applied with fraction = 0.08\%, DCPI improved the classification performance of ResNet-18 by 4.6\%. Additionally, for ResNet-101 with fraction = 3.90\%, DCPI yielded a 2.9\% performance gain, highlighting its robustness across different model architectures and dataset scales.

\begin{table*}[tb!]
\vspace{-5pt}
\caption{Results on ImageNet subsets when integrating DCPI into dataset distillation methods. Reduced datasets are initialized with MTT, and feature labels are synthesized with MTT.}
\vspace{-10pt}
\centering
\label{tab:mtt_subsets}
\resizebox{0.99\textwidth}{!}{
\begin{tabular}{@{}ccccccccccc@{}}
\toprule
Dataset &
  \multicolumn{2}{c}{ImageNette} &
  \multicolumn{2}{c}{ImageWoof} &
  \multicolumn{2}{c}{ImageFruit} &
  \multicolumn{2}{c}{ImageMeow} &
  \multicolumn{2}{c}{ImageYellow} \\ 
fraction(\%) &
  0.1 &
  1 &
  0.1 &
  1 &
  0.1 &
  1 &
  0.1 &
  1 &
  0.1 &
  1 \\ \midrule
MTT &
  47.7{\scriptsize$\pm$0.9} &
  63.0{\scriptsize$\pm$1.3} &
  28.6{\scriptsize$\pm$0.8} &
  35.8{\scriptsize$\pm$1.8} &
  26.6{\scriptsize$\pm$0.8} &
  40.3{\scriptsize$\pm$1.3} &
  30.7{\scriptsize$\pm$1.6} &
  40.4{\scriptsize$\pm$2.2} &
  45.2{\scriptsize$\pm$0.8} &
  60.0{\scriptsize$\pm$1.5} \\
\rowcolor{cyan!10}+DCPI &
  \textbf{50.5{\scriptsize$\pm$0.1}} &
  \textbf{65.7{\scriptsize$\pm$0.5}} &
  \textbf{31.3{\scriptsize$\pm$0.2}} &
  \textbf{37.5{\scriptsize$\pm$1.0}} &
  \textbf{29.1{\scriptsize$\pm$1.4}} &
  \textbf{43.0{\scriptsize$\pm$0.9}} &
  \textbf{34.0{\scriptsize$\pm$1.6}} &
  \textbf{43.8{\scriptsize$\pm$0.9}} &
  \textbf{46.6{\scriptsize$\pm$0.6}} &
  \textbf{62.2{\scriptsize$\pm$0.9}} \\
  \bottomrule
\end{tabular}
}
\end{table*}

\begin{table*}[tb!]
\centering
\caption{
Performance on ImageNet-1K when integrating DCPI into dataset distillation methods. Reduced datasets are initialized with RDED, and feature labels are synthesized with RDED. Data was distilled using both ResNet-18 and ConvNet, and the same models were employed for evaluation. In addition, we validated the dataset distilled by ResNet-18 using ResNet-101.
}
\vspace{-10pt}
\label{tab:rded}
\resizebox{0.99\textwidth}{!}{
\begin{tabular}{@{}c|ccc|ccc|cccccc@{}}
\toprule
\multirow{2.5}{*}{Fraction (\%)}    & \multicolumn{3}{c|}{ResNet-18} & \multicolumn{3}{c|}{ResNet-101} & \multicolumn{6}{c}{ConvNet} \\ 
\cmidrule(lr){2-13}
  & SRe2L & RDED & +DCPI  & SRe2L & RDED & +DCPI & MTT & IDM & TESLA & DATM & RDED &+DCPI \\  \midrule
  0.08   & 0.1{\scriptsize$\pm$0.1} & 6.6{\scriptsize$\pm$0.2} & \cellcolor{cyan!10}\textbf{11.2}{\scriptsize$\pm$0.6} & 0.6{\scriptsize$\pm$0.1} & 5.9{\scriptsize$\pm$0.4} & \cellcolor{cyan!10}\textbf{6.4}{\scriptsize$\pm$0.4} & - & - & \textbf{7.7}{\scriptsize$\pm$0.2} & - & 6.8{\scriptsize$\pm$0.2} & 7.0{\scriptsize$\pm$0.3} \\
 0.78  & 21.3{\scriptsize$\pm$0.6} & 42.0{\scriptsize$\pm$0.1} & \cellcolor{cyan!10}\textbf{43.7}{\scriptsize$\pm$0.3} & 30.9{\scriptsize$\pm$0.1} & 48.3{\scriptsize$\pm$1.0} & \cellcolor{cyan!10}\textbf{50.9}{\scriptsize$\pm$0.2} & - & - & 17.8{\scriptsize$\pm$1.3} & - & 20.4{\scriptsize$\pm$0.1} &  \cellcolor{cyan!10}\textbf{21.3}{\scriptsize$\pm$0.9} \\
 3.90  & 46.8{\scriptsize$\pm$0.2} & 56.5{\scriptsize$\pm$0.1} & \cellcolor{cyan!10}\textbf{58.1}{\scriptsize$\pm$0.2} & 60.8{\scriptsize$\pm$0.5} & 61.2{\scriptsize$\pm$0.4} & \cellcolor{cyan!10}\textbf{64.1}{\scriptsize$\pm$0.7} & - & - & 27.9{\scriptsize$\pm$1.2} & - & 38.4{\scriptsize$\pm$0.2} &  \cellcolor{cyan!10}\textbf{39.6}{\scriptsize$\pm$0.4} \\ \bottomrule
\end{tabular}
}
\label{tb:main}
\vspace{-5pt}
\end{table*}

\begin{table*}[tb!]
\centering
\captionof{table}{Cross-architecture evaluation of coresets on unseen networks. Reduced datasets are initialized with different pruning methods on CIFAR-10 (0.2\%). Feature labels are learned with DC. We utilized ConvNet for synthesizing feature labels.} 
\label{tab:corset_cross}
\vspace{-5pt}
{\resizebox{0.75\textwidth}{!}{\begin{tabular}{ccccccc}
\toprule
  & LeNet & MLP & ResNet & VGG & ConvNet & AlexNet \\ 
\midrule
 Herding & 23.0{\scriptsize$\pm$1.3 } & 21.3{\scriptsize$\pm$0.4 } & 26.2{\scriptsize$\pm$0.9 } & 24.1{\scriptsize$\pm$0.7 } & 25.1{\scriptsize$\pm$0.5 } & 23.3{\scriptsize$\pm$1.3 } \\ 
 \rowcolor{cyan!10}+DCPI & \textbf{32.4{\scriptsize$\pm$1.9 }} & \textbf{30.4{\scriptsize$\pm$0.4 }} & \textbf{36.9{\scriptsize$\pm$1.0 }} & \textbf{36.4{\scriptsize$\pm$0.7 }} & \textbf{46.3{\scriptsize$\pm$0.6 }} & \textbf{33.2{\scriptsize$\pm$2.0 }} \\ 
 ↑ & \color[HTML]{18A6CD}{9.4} & \color[HTML]{18A6CD}{9.1} & \color[HTML]{18A6CD}{10.7} & \color[HTML]{18A6CD}{12.3} & \color[HTML]{18A6CD}{21.1} & \color[HTML]{18A6CD}{9.9} \\ 
\midrule
Forgetting & 25.5{\scriptsize$\pm$1.6 } & 23.8{\scriptsize$\pm$0.3 } & 24.7{\scriptsize$\pm$1.0 } & 21.0{\scriptsize$\pm$0.4 } & 24.0{\scriptsize$\pm$0.5 } & 25.5{\scriptsize$\pm$0.9 } \\ 
 \rowcolor{cyan!10}+DCPI & \textbf{35.0{\scriptsize$\pm$1.6 }} & \textbf{33.1{\scriptsize$\pm$0.5 }} & \textbf{38.1{\scriptsize$\pm$0.9 }} & \textbf{35.3{\scriptsize$\pm$0.6 }} & \textbf{47.1{\scriptsize$\pm$0.7 }} & \textbf{35.7{\scriptsize$\pm$1.1 }} \\ 
 ↑ & \color[HTML]{18A6CD}{9.5} & \color[HTML]{18A6CD}{9.3} & \color[HTML]{18A6CD}{13.4} & \color[HTML]{18A6CD}{14.3} & \color[HTML]{18A6CD}{23.1} & \color[HTML]{18A6CD}{10.2} \\ 
\midrule
k-Center & 21.6{\scriptsize$\pm$1.1 } & 19.7{\scriptsize$\pm$0.4 } & 24.0{\scriptsize$\pm$0.8 } & 21.5{\scriptsize$\pm$0.8 } & 23.1{\scriptsize$\pm$0.7 } & 21.3{\scriptsize$\pm$0.7 } \\ 
 \rowcolor{cyan!10}+DCPI & \textbf{34.4{\scriptsize$\pm$1.3 }} & \textbf{31.4{\scriptsize$\pm$0.3 }} & \textbf{36.0{\scriptsize$\pm$1.5 }} & \textbf{34.3{\scriptsize$\pm$0.6 }} & \textbf{46.5{\scriptsize$\pm$0.6 }} & \textbf{36.0{\scriptsize$\pm$1.6 }} \\ 
 ↑ & \color[HTML]{18A6CD}{12.8} & \color[HTML]{18A6CD}{11.7} & \color[HTML]{18A6CD}{12.0} & \color[HTML]{18A6CD}{12.8} & \color[HTML]{18A6CD}{23.4} & \color[HTML]{18A6CD}{14.7} \\ 
\bottomrule
\end{tabular}}}
\vspace{-15pt}
\end{table*}

\subsection{Cross-architecture generalization}
\label{sec:Cross-architecture generalization}

Cross-architecture evaluation is a critical step toward ensuring robust generalization across previously unseen architectures. We measured the quality of reduced dataset with privileged information on both pruning and distillation settings. 
Incorporating diverse privileged information lets us tailor datasets to specific needs, while an additional fully connected layer resolves shape misalignment and maximizes reduced data potential.

For pruning methods, we utilized ConvNet to synthesize feature labels for selected coresets, and benchmarked their performance across 6 distinct network architectures. As illustrated in Table~\ref{tab:corset_cross}, on the CIFAR-10 (fraction = 0.2\%), all 3 methods yielded performance gains exceeding 20\% on ConvNet. These methods consistently exhibit improvements exceeding 10\% in most cases.

For distillation methods, we applied DCPI to synthesize feature labels for reduced datasets initialized by DC and MTT. As detailed in  Table~\ref{tab:mtt_cross} presents the results of applying DCPI to MTT on CIFAR-10 (fraction = 0.02\%), showing significant gains, such as an 11.1\% improvement when training on ConvNet and evaluating on AlexNet. Table~\ref{tab:comparison_exp}, we conducted experiments on distilled dataset initialized by DC on CIFAR-10 (fraction = 0.2\%), and trained models on 6 distinct network architectures, and evaluated the model performance across them. Notably, DCPI achieved an 18.3\% performance improvement over the baseline when training on VGG and evaluating on ResNet.

In addition to the results presented in Table~\ref{tab:corset_cross} and~\ref{tab:comparison_exp}, which summarize the performance of utilizing the synthesized datasets from pruning and dataset distillation to initialize the feature labels, we provide additional results in Table~\ref{tab:mtt_cross} to evaluate the cross-architecture generalization by using both feature labels and attention labels on different reduced datasets.

\begin{table}[tb!]
\centering
\caption{Cross-architecture performance comparison using feature labels and channel attention labels for initialization and learning on reduced datasets. We utilized reduced datasets initialized by both the DC and MTT methods. DC was employed for learning both feature labels and attention labels, while MTT was used exclusively for attention label learning through average pooling.}

\label{tab:mtt_cross}
\resizebox{0.79\linewidth}{!}{ %
\begin{tabular}{@{}c|cccccc@{}}
\toprule
\rowcolor[HTML]{FFFFFF} 
\textbf{}                                       & LeNet      & MLP        & ResNet     & VGG        & ConvNet    & AlexNet    \\ \midrule
\rowcolor[HTML]{FFFFFF} 
\cellcolor[HTML]{FFFFFF}DC                      & 16.8{\scriptsize$\pm$5.2 } & 28.0{\scriptsize$\pm$0.6 } & 36.2{\scriptsize$\pm$1.6 } & 35.4{\scriptsize$\pm$0.6 } & 44.5{\scriptsize$\pm$0.5 } & 20.1{\scriptsize$\pm$4.5 } \\
\rowcolor[HTML]{FFFFFF} 
\cellcolor[HTML]{FFFFFF}+DCPI-F         & 24.7{\scriptsize$\pm$4.3 } & 32.3{\scriptsize$\pm$1.0 } & 37.5{\scriptsize$\pm$1.8 } & 35.8{\scriptsize$\pm$0.8 } & 47.1{\scriptsize$\pm$0.9 } & 27.3{\scriptsize$\pm$3.2 } \\
\rowcolor[HTML]{FFFFFF} 
\cellcolor[HTML]{FFFFFF}+DCPI-A & 23.7{\scriptsize$\pm$6.6 } & 27.9{\scriptsize$\pm$0.5 } & 37.7{\scriptsize$\pm$1.1 } & 35.6{\scriptsize$\pm$0.7 } & 45.6{\scriptsize$\pm$0.5 } & 25.3{\scriptsize$\pm$3.0 } \\  \midrule
\rowcolor[HTML]{FFFFFF} 
\cellcolor[HTML]{FFFFFF}MTT                     & 29.1{\scriptsize$\pm$1.5 } & 29.1{\scriptsize$\pm$0.4 } & 35.1{\scriptsize$\pm$1.1 } & 31.4{\scriptsize$\pm$0.9 } & 45.6{\scriptsize$\pm$0.7 } & 24.0{\scriptsize$\pm$0.8 }  \\ 
\rowcolor[HTML]{FFFFFF} 
\cellcolor[HTML]{FFFFFF}+DCPI-A& 31.6{\scriptsize$\pm$1.7 }  & 29.4{\scriptsize$\pm$0.5 } & 36.7{\scriptsize$\pm$5.2 } & 37.9{\scriptsize$\pm$2.8 } & 47.0{\scriptsize$\pm$0.7 } & 34.1{\scriptsize$\pm$4.4 } \\ 
\bottomrule
\end{tabular}
}
\end{table}

Table~\ref{tab:mtt_cross} illustrates the cross-architecture evaluations conducted for both dataset condensation (DC) and MTT across various network architectures. In these experiments, we propose DCPI-F (feature labels learned by DC) and its variant DCPI-A, which is further enhanced by pooling through spatial attention. Similarly, feature labels synthesized with MTT are also average pooled into channel attention labels. We utilized several networks for cross evaluation, including LeNet, MLP, ResNet, VGG, ConvNet, and AlexNet.

We provide examples to illustrate the operation. For example, the first layer feature of a depth-3 ConvNet can be used to supervise the learning of feature labels. In this case, a single feature label takes the form of $Ch \times H \times W$ (\emph{e.g.}, $128\times 16\times 16$), which is reduced to a $128\times 1\times 1$ channel attention label after average pooling. Attention labels provide a more efficient way for using privileged information.

Additionally, during the cross-architecture process, if the features obtained from the training and testing models differ in shape or dimensionality, we employ an additional fully connected (FC) layer to align the features. For example, if the feature representation from the training model has a shape of $128 \times 16 \times 16$ and the testing model’s feature representation is $64 \times 16 \times 16$, the FC layer reshapes the $128 \times 16 \times 16$ feature into the $64 \times 16 \times 16$ format. The FC takes the input feature from the source architecture and transforms it into a format compatible with the target architecture by adjusting the dimensionality of the feature space.

The table demonstrates that initializing the reduced dataset with feature labels assigned through intermediate features of pre-trained networks leads to significant performance improvements across all architectures. Moreover, further gains are observed when channel attention pooling is applied. By leveraging Eq.~(\ref{eq:loss}) to learn and update the feature labels, the reduced dataset consistently yields competitive or improved performance across different networks. This highlights the benefits of feature and attention labels in enhancing model generalization, with particularly notable improvements observed in ConvNet and AlexNet settings. Channel attention pooling further contributes to these gains, reinforcing the effectiveness of this approach across architectures.
\vspace{-10pt}

\begin{table}[tb!]
    \centering
    \vspace{-5pt}
    \caption{Cross-architecture evaluations of distilled datasets on unseen networks. Reduced datasets are initialized with DC on CIFAR-10 (fraction = 0.2\%). Feature labels are learned with DC.}
    \label{tab:comparison_exp}
    \resizebox{0.78\textwidth}{!}{ %
    \begin{tabular}{c|*{2}{c}|*{2}{c}|*{2}{c}}
        \toprule
        Train\textbackslash Test & \multicolumn{2}{c|}{LeNet} & \multicolumn{2}{c|}{ConvNet} & \multicolumn{2}{c}{ResNet} \\ 
        & DC & +DCPI  & DC & +DCPI  & DC & +DCPI  \\ \midrule
        LeNet & 23.3{\scriptsize$\pm$5.3 } & \textbf{28.8{\scriptsize$\pm$4.1 }} (\textcolor[HTML]{18A6CD}{5.5}) & 35.8{\scriptsize$\pm$0.6 } & \textbf{46.6{\scriptsize$\pm$0.7 }} (\textcolor[HTML]{18A6CD}{10.8}) & 29.7{\scriptsize$\pm$2.0 } & \textbf{36.1{\scriptsize$\pm$1.3 }} (\textcolor[HTML]{18A6CD}{6.4}) \\
        MLP & 28.0{\scriptsize$\pm$1.1 } & \textbf{28.5{\scriptsize$\pm$4.1 }} (\textcolor[HTML]{18A6CD}{0.5}) & 29.0{\scriptsize$\pm$1.1 } & \textbf{46.5{\scriptsize$\pm$0.8 }} (\textcolor[HTML]{18A6CD}{17.6}) & 21.8{\scriptsize$\pm$1.8 } & \textbf{36.5{\scriptsize$\pm$1.2 }} (\textcolor[HTML]{18A6CD}{14.7}) \\
        ResNet & 22.0{\scriptsize$\pm$1.5 } & \textbf{25.9{\scriptsize$\pm$2.6 }} (\textcolor[HTML]{18A6CD}{3.9}) & 36.2{\scriptsize$\pm$0.8 } & \textbf{45.9{\scriptsize$\pm$0.6 }} (\textcolor[HTML]{18A6CD}{9.7}) & 33.6{\scriptsize$\pm$1.3 } & \textbf{36.1{\scriptsize$\pm$1.8 }} (\textcolor[HTML]{18A6CD}{2.5}) \\
        VGG & 22.7{\scriptsize$\pm$2.4 } & \textbf{28.3{\scriptsize$\pm$1.3 }} (\textcolor[HTML]{18A6CD}{5.6}) & 33.5{\scriptsize$\pm$1.1 } & \textbf{46.4{\scriptsize$\pm$0.6 }} (\textcolor[HTML]{18A6CD}{12.9}) &17.5{\scriptsize$\pm$1.9 } & \textbf{35.8{\scriptsize$\pm$1.5 }} (\textcolor[HTML]{18A6CD}{18.3}) \\
        ConvNet & 16.8{\scriptsize$\pm$1.6 } & \textbf{24.7{\scriptsize$\pm$4.3 }} (\textcolor[HTML]{18A6CD}{7.9}) & 44.5{\scriptsize$\pm$0.9 } & \textbf{47.1{\scriptsize$\pm$0.9 }} (\textcolor[HTML]{18A6CD}{2.6})  & 36.2{\scriptsize$\pm$1.6 } & \textbf{37.5{\scriptsize$\pm$1.1 }} (\textcolor[HTML]{18A6CD}{1.3}) \\
        AlexNet & 32.2{\scriptsize$\pm$1.9 } & \textbf{34.6{\scriptsize$\pm$1.6 }} (\textcolor[HTML]{18A6CD}{2.4}) & 36.9{\scriptsize$\pm$0.9 } & \textbf{46.2{\scriptsize$\pm$0.6 }} (\textcolor[HTML]{18A6CD}{9.3}) & 31.5{\scriptsize$\pm$1.5 } & \textbf{36.1{\scriptsize$\pm$1.4 }} (\textcolor[HTML]{18A6CD}{4.6}) \\ \midrule
    \end{tabular}
    }
    \vspace{-15pt}
\end{table}
\section{Discussion}
\label{sec:Discussion}
\begin{figure}[tb!]
    \centering
\includegraphics[width=0.99\linewidth]{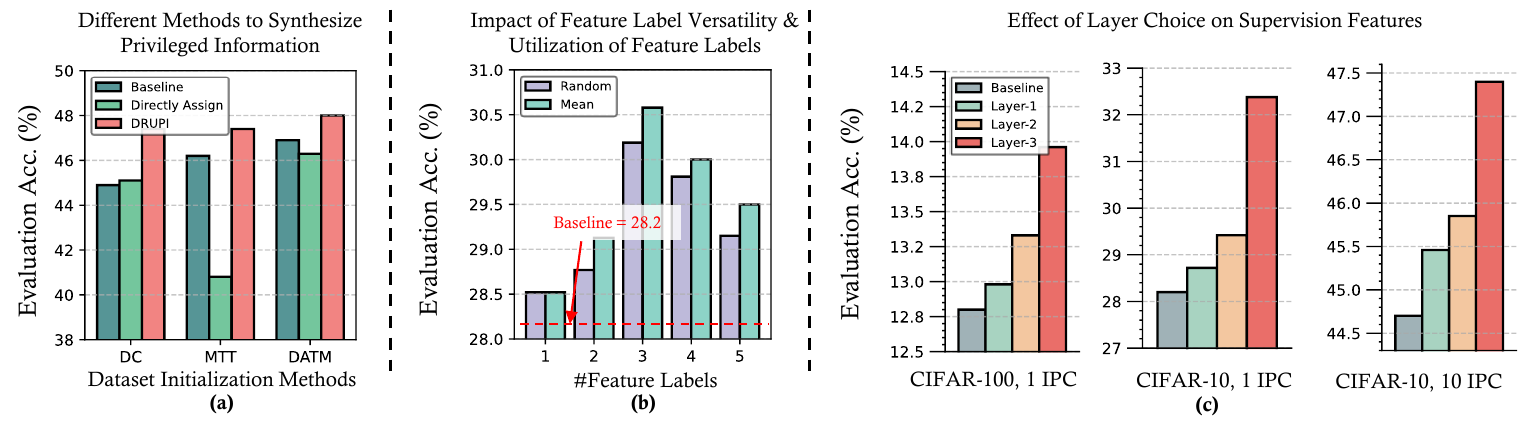}
    \vspace{-5pt}
    \caption{(a) Comparison of different methods for obtaining feature labels in datasets initialized with various distillation methods. Our results indicate that learning-based methods yield the best performance. (b) Impact of feature label versatility and the utilization of multiple feature labels. We find that incorporating more feature labels produces a more robust reduced dataset, with averaging the features outperforming random selection. (c) Evaluation of supervision using different layers from a depth-3 ConvNet for synthesizing feature labels. Results show that, across different IPCs and datasets, using the final layer features for supervision generates the most effective reduced dataset.}
    \label{fig:ablation}
    \vspace{-15pt}
\end{figure}

\vspace{-5pt}
\noindent \textbf{Different ways for synthesizing privileged information.} As discussed in Section~\ref{sec:Syn_PI}, there are multiple methods for synthesizing feature labels. A straightforward approach is to assign feature labels using a pre-trained model. However, as shown in Figure~\ref{fig:ablation}, this can sometimes degrade dataset quality. In contrast, learning feature labels offers greater flexibility and adaptability. As demonstrated in Figure~\ref{fig:ablation}(a), reduced datasets with learned feature labels significantly outperform those with directly assigned features. This is because feature labels extracted from a pre-trained model often lead to overly discriminative features with low diversity. The empirical results also support the observation that overly discriminative feature labels with strong task supervision can hurt performance in Figure~\ref{fig:ce_role}. More detailed results about the experiments on different methods are provided in Table~\ref{tab:cross_all} in Appendix~\ref{app:supervision_n_features}.

\noindent \textbf{Impact of feature label versatility and methods for utilizing feature labels.} We investigated the effect of synthesizing multiple feature labels for a single data-label pair. As shown in Figure~\ref{fig:ablation}(b), experiments  demonstrate that increasing the number of feature labels  enhances performance, likely due to the greater versatility captured by additional features. However, too many feature labels for a single input can introduce excessive diversity, leading to degraded performance. This verifies the trade-off between the diversity and discriminability of feature labels. Furthermore, averaging multiple feature labels outperforms random selection, which enables us to save only the averaged feature labels. Thus, increasing the number of feature labels introduces no extra storage overhead.

\noindent \textbf{Layer choice for supervision features.} We conducted an in-depth analysis to determine which ConvNet layer’s features are most effective for supervision. Specifically, we compared features extracted from the first, second, and final layers of a depth-3 ConvNet. Figure~\ref{fig:ablation}(c) shows that deeper layers consistently yielded better performance. This is likely due to the final layer’s ability to capture more complex and discriminative information, effectively representing high-level semantics. Therefore, we used the last layer’s features to supervise the synthesis of feature labels by default.

\noindent \textbf{Adaptive synthesis guidance applied in RDED.} When applied to RDED, DCPI operates as a flexible process guided by IPC. It leverages early-epoch, high-entropy soft labels\cite{nodistill} for scenarios with small IPC and late-epoch, full soft labels for scenarios with large IPC. For small IPC values, DCPI steers the data synthesis to either maximize diversity based on the early-stage teacher’s features (consistent with the high-entropy nature of its soft labels) or to amplify emerging discriminative signals, thereby enriching the limited data. In contrast, for large IPC settings, DCPI employs precise soft labels from fully converged teachers and prioritizes ensuring that the synthetic data demonstrates maximal discriminability within the converged teacher’s feature space, effectively converting label precision into highly informative and separable training examples.

For details on feature label initialization, see Appendix~\ref{app:init}. Ablation studies on MSE regression magnitude are in Table~\ref{tab:mse} and Appendix~\ref{app:reg}. Feature regression with other losses is explored in Table~\ref{tab:abla_main} and Appendix~\ref{app:ce_infonce}.
\vspace{-5pt}

\section{Conclusion}
\label{sec:Conclusion}

In this paper, we introduced DCPI, a novel framework that synthesizes privileged information for reduced datasets. To the best of our knowledge, DCPI is the first approach to go beyond the traditional data-label paradigm by utilizing synthesized feature labels. Extensive experiments on ImageNet, CIFAR-10/100 and Tiny ImageNet validate the effectiveness of DCPI, demonstrating significant improvements in model performance when integrated with existing condensation techniques. Additionally, we showed that achieving a balance between the discriminability and diversity of the synthesized feature labels is crucial for maximizing the quality of the reduced dataset.

\bibliography{colm2026_conference}
\bibliographystyle{colm2026_conference}

\appendix
\section{Detailed Experimental Settings}
\label{app:settings}

\subsection{Computational Resources}

The training was conducted on NVIDIA GPUs, specifically RTX 4090 and A100. All coreset selection experiments were run on A100 GPUs, while all DC-related experiments were carried out with RTX 4090. For MTT and DATM experiments on CIFAR-100 with fraction = 0.2\% and 2\%, as well as Tiny ImageNet, we utilized four NVIDIA A100 GPUs. Results of smaller datasets and lower fraction settings were conducted with RTX 4090.

\subsection{Coreset Selection}
\label{app:prune}
We first employed several coreset selection methods to initialize our reduced dataset, specifically using Herding~\cite{herding}, k-Center~\cite{k_center}, and Forgetting~\cite{forgetting}, where each data point was selected based on scores from a pre-trained ResNet-18. Next, we synthesized feature labels for the coreset using DC~\cite{DC}, assigning these labels to the intermediate features of a ConvNet trained for just one epoch. We also fine-tuned the initial images using the same method as for feature label learning, although we found that simply synthesizing feature labels could already bring performance improvement for reduced datasets.

Then we provide detailed settings for hyper-parameters. The hyper-parameters include:
\begin{enumerate}
    \item $\lambda_{reg}$: regularization coefficient, which controls the strength of the regularization term in the loss function.
    \item $\lambda_{task}$: task supervision coefficient, determining the discriminative power of the synthesized feature labels.
    \item $n_{feat}$: number of feature labels synthesized for a single data-label pair, which controls the diversity.
\end{enumerate}

Unless otherwise specified, for the CIFAR-10 dataset, we set $\lambda_{reg}$ to 0.5, while for CIFAR-100, it was set to 5. Across all configurations, the task supervision coefficient $\lambda_{task}$ was set to 0.001, and we only synthesized one feature label for a single data-label pair ($n_{feat}=1$) in the reduced dataset.

\subsection{Dataset Distillation}

\label{app:distillation}
For dataset distillation methods, we initialized the reduced datasets (images and labels) using distilled datasets and applied the same distillation method to synthesize feature labels. Specifically, we used DC~\cite{DC}, MTT~\cite{MTT}, and DATM~\cite{DATM} for both data-label initialization and feature label synthesis. 

We followed the original image-label synthesis settings of these distillation methods to generate feature labels. Table~\ref{tab:dd_setting_cifar_tiny} provide the detailed hyperparameter settings used for experiments on CIFAR-10, CIFAR-100, Tiny ImageNet. Table~\ref{tab:dd_setting_imagenet} provides the parameter settings for the MTT method on the ImageNet subsets. By default, we set $n_{feat}$ to 1 and $\lambda_{reg}$ to 0.01. We explored different configurations of fraction, specifically fraction(\%) = \{0.02, 0.2, 1\} for CIFAR-10, fraction = \{0.2, 2, 10\} for CIFAR-100 and fraction = \{0.2, 2\} for Tiny ImageNet.

\begin{table}[tb!]
    \centering
    \caption{Hyperparameter settings for dataset distillation methods. $^{\dagger}$ denotes that soft labels were synthesized in this set of experiments to further enrich the privileged information in the synthetic dataset. We used a pre-trained model to synthesize soft labels.}
    \label{tab:dd_setting_cifar_tiny}
    \resizebox{0.55\linewidth}{!}{
    \begin{tabular}{@{}c|c|c|c|c|c@{}}
    \toprule
    Dataset    & fraction(\%) & Method  & $\lambda_{reg}$ & $\lambda_{task}$ & $n_{feat}$ \\ \midrule
    \multirow{9}{*}{CIFAR-10}  & \multirow{3}{*}{0.02}   & DC    & 1.5        & 0.1          & 1        \\
                              &                       & MTT$^{\dagger}$   & 0.5        & 0.001        & 1        \\
                              &                       & DATM  & 0.5        & 0.001        & 5        \\ \cmidrule(lr){2-6}
                              & \multirow{3}{*}{0.2}  & DC    & 0.5        & 0.1          & 1        \\
                              &                       & MTT   & 0.0005     & 0.01         & 3        \\
                              &                       & DATM  & 0.05       & 0.1          & 3        \\ \cmidrule(lr){2-6}
                              & \multirow{3}{*}{1}  & DC    & 0.01       & 0.01         & 1        \\
                              &                       & MTT   & 0.05       & 0.001        & 1        \\
                              &                       & DATM  & 0.05       & 0.001        & 1        \\ \midrule
    \multirow{9}{*}{CIFAR-100} & \multirow{3}{*}{0.2}   & DC    & 1.5        & 0.1          & 1        \\
                              &                       & MTT   & 0.5        & 0.01         & 3        \\
                              &                       & DATM  & 0.05       & 0.01         & 1        \\ \cmidrule(lr){2-6}
                              & \multirow{3}{*}{2}  & DC$^{\dagger}$    & 0.001      & 0.005        & 1        \\
                              &                       & MTT   & 0.005      & 0.001        & 1        \\
                              &                       & DATM  & 0.05       & 0.001        & 1        \\ \cmidrule(lr){2-6}
                              & \multirow{3}{*}{10}  & DC$^{\dagger}$    & 0.5        & 0.1          & 1        \\
                              &                       & MTT$^{\dagger}$   & 0.5        & 0.01         & 3        \\
                              &                       & DATM  & 0.0005     & 0.001        & 1        \\ \midrule
    \multirow{5}{*}{Tiny ImageNet} & \multirow{2}{*}{0.2}   & MTT$^{\dagger}$   & 0.5        & 0.0001       & 3        \\
                              &                       & DATM  & 0.005      & 0.001        & 1        \\ \cmidrule(lr){2-6}
                              & \multirow{2}{*}{2}  & MTT$^{\dagger}$   & 0.005      & 0.001        & 1        \\
                              &                       & DATM  & 0.005      & 0.001        & 1        \\ \bottomrule
    \end{tabular}
    }
    \vspace{-10pt}
\end{table}
 
\begin{table}[tb!]
    \centering
    \caption{Hyperparameter settings for ImageNet subsets used in MTT experiments.}
    \label{tab:dd_setting_imagenet}
    \resizebox{0.30\linewidth}{!}{
    \begin{tabular}{@{}c|c|c@{}}
    \toprule
    Dataset      & fraction & $\lambda_{reg}$ \\ \midrule
    \multirow{2}{*}{ImageNette}  
                 & 0.1   &  0.005               \\ 
                 & 1  &  0.005               \\ \midrule
    \multirow{2}{*}{ImageWoof} 
                 & 0.1   &  0.005               \\ 
                 & 1  &  0.5               \\ \midrule
    \multirow{2}{*}{ImageFruit} 
                 & 0.1   &  0.005        \\ 
                 & 1  &  0.5               \\ \midrule
    \multirow{2}{*}{ImageMeow} 
                 & 0.1   &  0.00005             \\ 
                 & 1  &  0.5               \\ \midrule
    \multirow{2}{*}{ImageYellow} 
                 & 0.1   &  0.005            \\ 
                 & 1  &  0.5             \\ \bottomrule
    \end{tabular}
    }
    \vspace{-5pt}
\end{table}
For each method, the hyperparameters for feature synthesis are fine-tuned across different datasets and fraction settings to achieve optimal performance. For instance, in CIFAR-10 (fraction = 0.02\%), the DC method utilizes $\lambda_{reg} = 1.5$, $\lambda_{task} = 0.1$, and $n_{feat} = 1$. In contrast, for fraction = 1\%, the same method adjusts its hyperparameters to $\lambda_{reg} = 0.01$, $\lambda_{task} = 0.01$, and $n_{feat} = 1$. Similar fine-tuning is performed for all datasets and fraction values across each method.

For distilled datasets with feature labels, we also tried to incorporate additional forms of privileged information, such as soft labels, to further enrich the privileged information in the synthetic dataset. Specifically, we used a pre-trained network to generate soft labels for the distilled dataset. Some methods like DATM have already learned soft labels. We only synthesized soft labels for DC and MTT based reduced datasets. We provide futher results on soft labels in Table~\ref{tab:abla_main}.

\section{Additional Results on DCPI}

\subsection{Further Performance Results}
\label{app:further_results}

In Section~\ref{sec:Main_results}, we presented results for two dataset distillation methods, DC and MTT. Here, we provide the results for DATM on the CIFAR-10 and CIFAR-100 datasets. We initialized the images and labels with DATM and used it to synthesize feature labels for the reduced dataset. Experiments were conducted using a ConvNet for both distillation and evaluation tasks. We additionally provide baselines such as \cite{FRePo,RCIG,TESLA,FTD,DATM} .Table~\ref{tab:datm_main} shows the performance improvements of DCPI over various existing methods under different fraction settings. Specifically, the datasets used in these experiments include CIFAR-10 with fraction(\%) = \{0.02, 0.2, 1\} and CIFAR-100 with fraction(\%) = \{0.2, 2\}. Each dataset was further evaluated with varying data fractions, allowing us to assess the generalization of the methods across different data availability scenarios.

\begin{table}[tb!]
\centering
\caption{The application of DCPI to DATM across CIFAR-10, CIFAR-100, accompanied by a comparative analysis with existing methods. ConvNet is utilized for both distillation and evaluation. Our methodology demonstrates enhanced performance compared to previous results.The ↑ symbol signifies performance enhancements compared to random selection.}
\label{tab:datm_main}
\resizebox{0.7\linewidth}{!}{  
\begin{tabular}{c|*{3}{c}|c*{1}{c}}
\toprule
 Dataset & \multicolumn{3}{c|}{CIFAR-10} & \multicolumn{2}{c}{CIFAR-100} \\ 
Fraction (\%) & 0.02 & 0.2 & 1 & 0.2 & 2 \\\midrule
Random & 15.4{\scriptsize$\pm$0.3 } & 31.0{\scriptsize$\pm$0.5 } & 50.6{\scriptsize$\pm$0.3 } & 4.2{\scriptsize$\pm$0.3 } & 14.6{\scriptsize$\pm$0.5 } \\ 
KIP & 49.9{\scriptsize$\pm$0.2 } & 62.7{\scriptsize$\pm$0.3 } & 68.6{\scriptsize$\pm$0.3 } & 15.7{\scriptsize$\pm$0.2 } & 28.3{\scriptsize$\pm$0.1 } \\ 
FRePo & 46.8{\scriptsize$\pm$0.7 } & 65.5{\scriptsize$\pm$0.4 } & 71.7{\scriptsize$\pm$0.4 } & 28.7{\scriptsize$\pm$0.2 } & 42.5{\scriptsize$\pm$0.4 } \\ 
RCIG & 53.9{\scriptsize$\pm$1.0 } & 69.1{\scriptsize$\pm$0.4 } & 73.5{\scriptsize$\pm$0.3 } & 39.3{\scriptsize$\pm$0.4 } & 44.1{\scriptsize$\pm$0.4 } \\ 
DM & 26.0{\scriptsize$\pm$0.8 } & 48.9{\scriptsize$\pm$0.6 } & 63.0{\scriptsize$\pm$0.4 } & 11.4{\scriptsize$\pm$0.3 } & 29.7{\scriptsize$\pm$0.3 } \\ 
DSA & 28.8{\scriptsize$\pm$0.7 } & 52.1{\scriptsize$\pm$0.5 } & 60.6{\scriptsize$\pm$0.5 } & 13.9{\scriptsize$\pm$0.3 } & 32.3{\scriptsize$\pm$0.3 } \\ 
DCC & 32.9{\scriptsize$\pm$0.8 } & 49.4{\scriptsize$\pm$0.5 } & 61.6{\scriptsize$\pm$0.4 } & 13.3{\scriptsize$\pm$0.3 } & 30.6{\scriptsize$\pm$0.4 } \\ 
DSAC & 34.0{\scriptsize$\pm$0.7 } & 54.5{\scriptsize$\pm$0.5 } & 64.2{\scriptsize$\pm$0.4 } & 14.6{\scriptsize$\pm$0.3 } & 33.5{\scriptsize$\pm$0.3 } \\ 
CAFE & 30.3{\scriptsize$\pm$1.1 } & 46.3{\scriptsize$\pm$0.6 } & 55.5{\scriptsize$\pm$0.6 } & 12.9{\scriptsize$\pm$0.3 } & 27.8{\scriptsize$\pm$0.3 } \\ 
IDM & 45.6{\scriptsize$\pm$0.7 } & 58.6{\scriptsize$\pm$0.1 } & 67.5{\scriptsize$\pm$0.1 } & 20.1{\scriptsize$\pm$0.3 } & 45.1{\scriptsize$\pm$0.1 } \\ 
TESLA & \textbf{48.5{\scriptsize$\pm$0.8 }} & 66.4{\scriptsize$\pm$0.8 } & 72.6{\scriptsize$\pm$0.7 } & 24.8{\scriptsize$\pm$0.4 } & 41.7{\scriptsize$\pm$0.3 } \\ 
FTD & 46.0{\scriptsize$\pm$0.4 } & 65.3{\scriptsize$\pm$0.4 } & 73.2{\scriptsize$\pm$0.2 } & 24.4{\scriptsize$\pm$0.4 } & 42.5{\scriptsize$\pm$0.2 } \\ 
\midrule
DATM & 46.9{\scriptsize$\pm$0.5 } & 66.8{\scriptsize$\pm$0.2 } & 76.1{\scriptsize$\pm$0.3 } & 27.9{\scriptsize$\pm$0.2 } & 47.2{\scriptsize$\pm$0.4 } \\ 
\rowcolor{cyan!10}+DCPI & 48.0{\scriptsize$\pm$0.4 } & \textbf{68.08{\scriptsize$\pm$0.1 }} & \textbf{76.2{\scriptsize$\pm$0.06 }} & \textbf{28.4{\scriptsize$\pm$0.5 }} & \textbf{55.21{\scriptsize$\pm$0.06 }} \\  
↑ &{\color[HTML]{18A6CD} 1.1} & {\color[HTML]{18A6CD} 1.0} & {\color[HTML]{18A6CD} 0.3} & {\color[HTML]{18A6CD} 0.5} & {\color[HTML]{18A6CD} 0.4} \\ 
\midrule
 Full Dataset & \multicolumn{3}{c|}{84.8{\scriptsize$\pm$0.1 }} & \multicolumn{2}{c}{56.2{\scriptsize$\pm$0.3 }} \\ 
 
\bottomrule
\end{tabular}
}
\end{table}

\subsection{Additional Results on Different Methods for Synthesizing Privileged Information}
\label{app:supervision_n_features}

As a complement to Figure~\ref{fig:ablation}(a), Table~\ref{tab:cross_all} provides a more comprehensive overview of the experimental results on different methods for synthesizing privileged information. Specifically, it presents a detailed comparison between direct feature label assignment and the DCPI framework across three dataset distillation methods: DC, MTT, and DATM. These evaluations were conducted on CIFAR-10, CIFAR-100, and Tiny ImageNet datasets with varying fraction values, and include results for different fractions of the full dataset.

For each method, the results are provided for various fraction settings, showing the performance both with direct feature assignment and with the inclusion of DCPI (feature label). The DCPI framework consistently improves performance across different methods and datasets, as indicated by the higher accuracy values. Specifically, DCPI shows substantial improvements over direct assignment, particularly in the low-data regime, such as CIFAR-100(0.2\%) and Tiny ImageNet(0.2\%). These findings underscore the effectiveness of DCPI in enhancing the generalization and performance of dataset distillation techniques, even with limited data.

\begin{table}[tb!]
\centering

\caption{Experimental results comparing direct feature label assignment and the DCPI framework across three dataset distillation methods (DC, MTT, DATM) on various datasets with different fraction values. }
\label{tab:cross_all}
\resizebox{0.99\linewidth}{!}{
\begin{tabular}{c|*{3}{c}|c*{2}{c}|c}
\toprule
 Dataset & \multicolumn{3}{c|}{CIFAR-10} & \multicolumn{3}{c|}{CIFAR-100} & Tiny ImageNet \\ 
Fraction (\%) & 0.02 & 0.2 & 1 & 0.2 & 2 & 10 & 0.2  \\\midrule
DC & 28.3{\scriptsize$\pm$0.5 } & 44.9{\scriptsize$\pm$0.5 } & 53.9{\scriptsize$\pm$0.5 } & 12.8{\scriptsize$\pm$0.3 } & 25.2{\scriptsize$\pm$0.3 } & 29.8{\scriptsize$\pm$0.3 } & - \\ 
\rowcolor{gray!10} Directly Assign & 28.3{\scriptsize$\pm$0.8 } & 45.1{\scriptsize$\pm$0.5 } & 54.1{\scriptsize$\pm$0.4 } & 12.7{\scriptsize$\pm$0.4 } & 25.1{\scriptsize$\pm$0.4 }& 29.8{\scriptsize$\pm$0.4 } & -  \\
\rowcolor{cyan!10} DCPI & 31.5{\scriptsize$\pm$0.9 } & 47.3{\scriptsize$\pm$0.9 } & 55.3{\scriptsize$\pm$0.5 } & 14.9{\scriptsize$\pm$0.4 } & 25.6{\scriptsize$\pm$0.5 }& 31.6{\scriptsize$\pm$0.5 } & -  \\
\midrule
MTT & 46.2{\scriptsize$\pm$0.8 } & 65.4{\scriptsize$\pm$0.7 } & 71.6{\scriptsize$\pm$0.2 } & 24.3{\scriptsize$\pm$0.3 } & 39.7{\scriptsize$\pm$0.4 } & 47.7{\scriptsize$\pm$0.2 } & 8.8{\scriptsize$\pm$0.3 } \\ 
\rowcolor{gray!10} Directly Assign & 40.8{\scriptsize$\pm$1.8 } & 56.2{\scriptsize$\pm$1.1 } & 66.1{\scriptsize$\pm$0.5 } & 23.9{\scriptsize$\pm$0.5 } & 38.1{\scriptsize$\pm$0.4 }& 47.4{\scriptsize$\pm$0.2 } & 8.1{\scriptsize$\pm$0.4 }  \\
\rowcolor{cyan!10} DCPI & 47.4{\scriptsize$\pm$0.5 } & 65.9{\scriptsize$\pm$0.6 } & 72.7{\scriptsize$\pm$0.2 } & 25.6{\scriptsize$\pm$0.4 } & 41.0{\scriptsize$\pm$0.3 } & 48.4{\scriptsize$\pm$0.3 } & 11.2{\scriptsize$\pm$0.1 } \\
\midrule
DATM & 46.9{\scriptsize$\pm$0.5 } & 66.8{\scriptsize$\pm$0.2 } & 76.1{\scriptsize$\pm$0.3 } & 27.9{\scriptsize$\pm$0.2 } & 47.2{\scriptsize$\pm$0.4 } & 55.0{\scriptsize$\pm$0.2 } & 17.1{\scriptsize$\pm$0.3 } \\ 
\rowcolor{gray!10} Directly Assign & 46.3{\scriptsize$\pm$0.7 } & 64.5{\scriptsize$\pm$0.5 } & 73.7{\scriptsize$\pm$0.4 } & 26.6{\scriptsize$\pm$0.3 } & 36.2{\scriptsize$\pm$0.5 } & 55.5{\scriptsize$\pm$0.3 } & 15.6{\scriptsize$\pm$0.1 }  \\
\rowcolor{cyan!10} DCPI & 48.0{\scriptsize$\pm$0.4 } & 68.1{\scriptsize$\pm$0.3 } & 76.2{\scriptsize$\pm$0.3 } & 28.4{\scriptsize$\pm$0.5 } & 47.6{\scriptsize$\pm$0.2 } & 55.2{\scriptsize$\pm$0.1 } & 17.6{\scriptsize$\pm$0.1 } \\  
\midrule
 Full Dataset & \multicolumn{3}{c|}{84.8{\scriptsize$\pm$0.1 }} & \multicolumn{3}{c|}{56.2{\scriptsize$\pm$0.3 }} & 37.6{\scriptsize$\pm$0.4 } \\ 
\bottomrule
\end{tabular}
}
\vspace{-5pt}
\end{table}

\subsection{Effects on Feature Label Initialization}
\label{app:init}

We further explored the issue of feature label initialization. Specifically, initialization can be performed either by using random noise or by feeding synthetic images into a ConvNet to extract intermediate layer features for initialization.

\begin{figure}[htbp]
    \centering
    \vspace{-15pt}
    \includegraphics[width=0.89\linewidth]{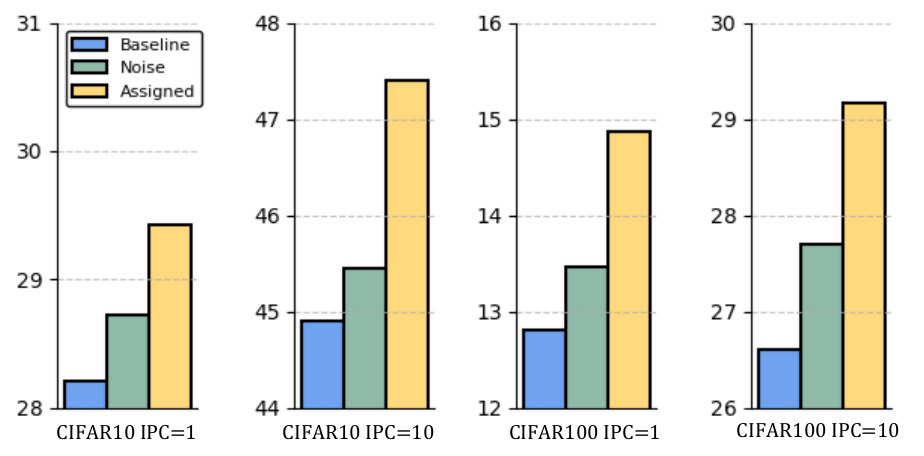}

    \caption{Comparison of noise initialization (yellow) and initialization with assigned features (blue) from a pre-trained ConvNet on CIFAR-10 and CIFAR-100 across different fraction settings.}
    \label{fig:init}
    \vspace{-20pt}
\end{figure}

Figure~\ref{fig:init} presents a performance comparison between two feature initialization approaches—random noise and assigned features—across the CIFAR-10 and CIFAR-100 datasets with varying fraction settings. In this comparison, noise (yellow bars) refers to randomly initialized feature labels, while real (blue bars) refers to initialization using features extracted from synthetic datasets passed through the network.

For CIFAR-10 (IPC = 1) setting, the initialization with assigned features method achieves significantly higher performance (around 33\%) compared to the noise-based initialization (approximately 30\%). A similar trend is observed in the CIFAR-10 (IPC = 10) case, where initialization with assigned features substantially outperforms noise.

On the CIFAR-100 (IPC = 1) dataset, initialization with assigned features also demonstrates better performance, with a clear margin over noise initialization, where using assigned features reaches around 29\% while noise remains below 28\%. The same pattern holds for CIFAR-100 (IPC = 10), where initialization with assigned features achieves a accruacy of over 47\%, far exceeding the noise initialization's performance of approximately 45\%.

\section{Ablation on Regression Supervision}
\label{app:supervision}

\subsection{Sensitivity of Regression Magnitude}
\label{app:reg}

We further investigated the influence of $\lambda_{reg}$ on the model's performance. The results in Table~\ref{tab:mse} demonstrate that varying the MSE regularization parameter $\lambda_{reg}$ does not significantly impact the final accuracy. While there is a slight improvement as $\lambda_{reg}$ increases, from 28.91\% at $\lambda_{reg} = 0.05$ to 30.74\% at $\lambda_{reg} = 10$, the overall effect remains relatively small. This suggests that the model's performance is not highly sensitive to the regularization weight for MSE loss within the tested range, indicating that other factors may have a more dominant influence on accuracy. In this case, regularization helps prevent overfitting but does not drastically change the model’s ability to generalize within the fraction = 0.02\% setting for CIFAR-10.

\begin{table}[tb!]
  \centering
  \caption{Results on CIFAR-10 (fraction = 0.02\%) for different $\lambda_{reg}$.}
  \label{tab:mse}
  \vspace{-0.3em}  
  \resizebox{0.54\linewidth}{!}{
    \begin{tabular}{ccccccc}
      \toprule
      $\lambda_{reg}$ & 0.05 & 0.1 & 0.5 & 1 & 5 & 10 \\
      \midrule
      Acc (\%) & 28.91 & 29.20 & 30.42 & 30.53 & 30.67 & 30.74 \\
      \bottomrule
    \end{tabular}
  }
\end{table}

\subsection{Further Regression Supervision Objectives}
\label{app:ce_infonce}
We extended the use of feature labels by integrating additional supervision mechanisms, including CE loss and InfoNCE loss, to enhance the distillation process. Additionally, we performed ablation studies to evaluate the impact of using soft labels. These components contribute to a more structured and informative representation learning framework.

First, we introduced CE loss and InfoNCE loss to provide direct regularization between the feature labels and the supervision features. In line with previous work, we also incorporated soft labels as privileged information.

The combination of CE loss, InfoNCE loss, and soft labels consistently yielded the best performance across our ablation studies, as shown in Tables~\ref{tab:abla_main}. We found that incorporating InfoNCE or CE loss marginally improved the quality of the reduced dataset, while soft labels, when combined with feature labels, provided a more significant boost in performance.

\begin{table}[t]
  \centering
  \caption{Ablation of DCPI components on CIFAR-10 with IPC = 1 and CIFAR-100 with IPC = 1. The table compares the effects of feature labels, CE loss ($\mathcal{L}_{CE}$), contrastive loss ($\mathcal{L}_{IN}$), and soft labels on model performance.}
  \label{tab:abla_main}
  \vspace{-0.35em}

  \setlength{\tabcolsep}{4pt}
  \renewcommand{\arraystretch}{1.05}
  \scriptsize

  \begin{subtable}[t]{0.48\linewidth}
    \centering
    \resizebox{\linewidth}{!}{
    \begin{tabular}{ccccc}
      \toprule
      Feature & $\mathcal{L}_{CE}$ & $\mathcal{L}_{IN}$ & Soft & Acc (\%) \\
      \midrule
       &  &  &  & 28.30 \\
      \checkmark &  &  &  & 31.13 \\
      \checkmark & \checkmark &  &  & 31.54 \\
      \checkmark &  & \checkmark &  & 31.47 \\
      \checkmark & \checkmark & \checkmark &  & 31.90 \\
      \checkmark &  &  & \checkmark & 32.28 \\
      \bottomrule
    \end{tabular}}
    \caption{CIFAR-10, fraction = 0.02\%}
    \label{tab:abla_c10ipc1}
  \end{subtable}\hfill
  \begin{subtable}[t]{0.48\linewidth}
    \centering
    \resizebox{\linewidth}{!}{
    \begin{tabular}{ccccc}
      \toprule
      Feature & $\mathcal{L}_{CE}$ & $\mathcal{L}_{IN}$ & Soft & Acc (\%) \\
      \midrule
       &  &  &  & 12.80 \\
      \checkmark &  &  &  & 13.96 \\
      \checkmark & \checkmark &  &  & 14.10 \\
      \checkmark &  & \checkmark &  & 14.38 \\
      \checkmark & \checkmark & \checkmark &  & 14.64 \\
      \checkmark &  &  & \checkmark & 14.86 \\
      \bottomrule
    \end{tabular}}
    \caption{CIFAR-100, fraction = 0.2\%}
    \label{tab:abla_c100ipc1}
  \end{subtable}

  \vspace{-3em}
\end{table}

\section{Pseudo Code of DCPI}
\label{app:pseudo_code}

Algorithm~\ref{alg:process} outlines the process where we initialize the reduced dataset with assigned feature labels based on the synthetic dataset from dataset condensation (DC). Subsequently, we update the reduced dataset by learning the feature labels, which are progressively refined during the training process.

\begin{algorithm}
\caption{Dataset Condensation using Privileged Information (DCPI) For DC}
\label{alg:process}
\begin{algorithmic}[1]

\Require  Outer-loop steps $K$, inner-loop steps $T$,  synthesized privileged information $\{f_i^\star\}_{i=1}^m$. Let $\psi(\cdot)$ denote the intermediate output of model $g$, and $g = \psi \circ \kappa$, where $\kappa(\cdot)$ is the classifier component of $g$. And $\sigma(\cdot)$ represents the softmax function.

\State Initialize reduced dataset $\mathcal{D}_{\mathcal{S}} = \{(\tilde{x}_i, \tilde{y}_i)\}_{i=1}^m$ and privileged information. The initial feature labels are derived from a pre-trained network $g_\mathcal{T}$ applied to the synthetic dataset.

\For{$k = 0, \dots, K-1$}
    \State Initialize neural network $g$'s weights $\theta_0$
    \For{$t = 0, \dots, T-1$}
        \For{$c = 0, \dots, C-1$}
            \State Sample mini-batches from $\mathcal{\mathcal{D}_{\mathcal{T}}}$ and $\mathcal{D}_{\mathcal{S}}$
            \State Compute loss the update synthetic privileged information  using  Eq.~(\ref{eq:loss})
        \EndFor
    \EndFor
\EndFor
\State \textbf{Output:} An  Optimized and extended dataset represented as $\mathcal{D}_{\mathcal{S}}^\star = {(\tilde{x}_i, \tilde{y}_i, f_i^\star)}$
\end{algorithmic}

\end{algorithm}

\subsection{Theoretical Analysis}
\label{sec:Why_PI}
We analyze how the DCPI improves reduced dataset quality. Let $g_{\mathcal{T}} \in \mathcal{G}_{\mathcal{T}}$ be the oracle function for dataset $\mathcal{D}_{\mathcal{T}}$, and $|\cdot|_C$ denote a model performance measure. Consider two models: $g_{\mathcal{S}} \in \mathcal{G}_{\mathcal{S}}$ trained on reduced dataset $\mathcal{D}_{\mathcal{S}}$, and $g_{\mathcal{S}^\star} \in \mathcal{G}_{\mathcal{S}^\star}$ trained on $\mathcal{D}_{\mathcal{S}^\star}$ with privileged information.

Using VC theory~\cite{vc_theory}, we assess its performance via VC-dimension. For a model $g \in \mathcal{G}$ with finite VC-dimension $|\mathcal{G}|_{\text{VC}}$, the expected error $R(g)$ is bounded with probability $1-\delta$ as:
\begin{equation}
    R(g) \leq R_m(g)+O\left(\left(\frac{|\mathcal{G}|_{\mathrm{VC}}-\log \delta}{m}\right)^\alpha\right),
\end{equation}
where $O(\cdot)$ is the estimation error, $R_m(g)$ is the training error over $m$ data points, and $\alpha \in [\frac{1}{2}, 1]$ reflects task difficulty. For non-separable tasks, $\alpha \approx \frac{1}{2}$, giving a learning rate of $O(m^{-1/2})$. For separable tasks with no training errors, $\alpha \approx 1$, yielding $O(m^{-1})$. A teacher model can accelerate learning from $O(m^{-1/2})$ to $O(m^{-1})$ for a student on a dataset of size $m$.

We analyze DCPI, showing how privileged information speeds up learning and enhances reduced dataset quality. Extending~\cite{unify_privileged_info}, we apply results to dataset condensation. The model $g_{\mathcal{S}}$ on reduced dataset $\mathcal{D}_{\mathcal{S}}$ learns the true function $g_{\mathcal{T}}$ at a slower rate $\alpha_{\mathcal{S}}$:
\begin{equation}
\label{eq:syn_real}
  R(g_{\mathcal{S}}) - R(g_{\mathcal{T}}) \leq O\left(\frac{|\mathcal{G}_{\mathcal{S}}|_{\text{C}}}{m^{\alpha_{\mathcal{S}}}}\right) + \varepsilon_{\mathcal{S}},
\end{equation}
where $\varepsilon_{\mathcal{S}}$ is the approximation error of $\mathcal{G}_{\mathcal{S}}$ with respect to $g_{\mathcal{T}} \in \mathcal{G}_{\mathcal{T}}$. Second, assume that the model trained on the dataset with privileged information $g_{\mathcal{S}^\star}$ learns at a faster rate $\alpha_{\mathcal{S}^\star}$:
\begin{equation}
\label{eq:PI_real}
  R(g_{\mathcal{S}^\star}) - R(g_{\mathcal{T}}) \leq O\left(\frac{|\mathcal{G}_{\mathcal{S}^\star}|_{\text{C}}}{m^{\alpha_{\mathcal{S}^\star}}}\right) + \varepsilon_{\mathcal{S}^\star},
\end{equation}
where $\varepsilon_{\mathcal{S}^\star}$ is the approximation error of $\mathcal{G}_{\mathcal{S}^\star}$ with respect to $g_{\mathcal{T}}$. Finally, assume that the performance difference $g_{\mathcal{S}}$ learns from the model with privileged information $g_{\mathcal{S^\star}}$ is
\begin{equation}
\label{eq:PI_syn}
  R(g_{\mathcal{S}}) - R(g_{\mathcal{S}^\star}) \leq O\left(\frac{|\mathcal{G}_{\mathcal{S}}|_{\text{C}}}{m^{\alpha}}\right) + \varepsilon,
\end{equation}
where $\varepsilon$ is the approximation error of $\mathcal{G}_{\mathcal{S}}$ with respect to $g_{\mathcal{S}^\star}$, and $\frac{1}{2} \leq \alpha \leq 1$. Combining Eq.~(\ref{eq:PI_real}) and Eq.~(\ref{eq:PI_syn}), the learning rate for the model without privileged information learning the oracle function $g_{\mathcal{T}}$ is then given by
\begin{equation}
\begin{aligned}
  & R(g_{\mathcal{S}})-R(g_{\mathcal{T}}) = R(g_{\mathcal{S}})-R(g_{\mathcal{S^\star}})+R(g_{\mathcal{S^\star}})-R(g_{\mathcal{T}})\\
                  &\leq
                  O\left(\frac{|\mathcal{G}_{\mathcal{S}}|_\textrm{C}}{m^{\alpha}}\right)
                  + \varepsilon +
                  O\left(\frac{|\mathcal{G}_{\mathcal{S^\star}}|_\textrm{C}}{m^{\alpha_{\mathcal{S}^\star}}}\right) +
                  \varepsilon_{\mathcal{S^\star}}
                  \leq O\left(\frac{|\mathcal{G}_{\mathcal{S}}|_\textrm{C}}{m^{\alpha}}+\frac{|\mathcal{G}_{\mathcal{S^\star}}|_\textrm{C}}{m^{\alpha_{\mathcal{S}^\star}}}\right)
                  + \varepsilon  +
                  \varepsilon_{\mathcal{S^\star}},
\end{aligned}
\end{equation}
where the final inequality arises because $\alpha \leq 1$. Therefore, for $\frac{1}{2} < \alpha \leq 1$ in dataset condensation settings, the inequality becomes:
\begin{equation}
O\left(\frac{|\mathcal{G}_{\mathcal{S}}|_\textrm{C}}{m^{\alpha}}+\frac{|\mathcal{G}_{\mathcal{S^\star}}|_\textrm{C}}{m^{\alpha_{\mathcal{S}^\star}}}\right) + \varepsilon + \varepsilon_{\mathcal{S}^\star} \leq O\left(\frac{|\mathcal{G}_{\mathcal{S}}|_{\text{C}}}{{m}^{\alpha_\mathcal{S}}}\right) + \varepsilon_{\mathcal{S}}.
\end{equation}
This inequality highlights the advantages of models trained with privileged information: training using privileged information exhibit lower generalization and approximation errors compared to those trained without privileged information. More importantly, it emphasizes that the privileged information is most beneficial in low-data regimes, which is the typical DR scenario. These benefits align with the principles of LUPI as outlined in ~\cite{vapnik2015learning,unify_privileged_info}.

\section{Further Discussion about DCPI}

\label{sec:impact}
The advancement of dataset condensation techniques like DCPI holds significant implications for the broader machine learning landscape. By enabling the creation of smaller, yet more statistically potent datasets, DCPI can democratize access to training sophisticated models, particularly for entities with constrained computational resources. This could accelerate research and development in various domains by lowering the barrier to entry for working with large-scale data. The ability to synthesize richer supervisory signals beyond simple data-label pairs may also foster the development of more data-efficient learning algorithms and contribute to models with enhanced generalization capabilities. From a sustainability perspective, training on compact datasets naturally translates to reduced energy consumption and a smaller carbon footprint associated with model development cycles. However, it is crucial to consider the potential for bias amplification. If the original large dataset, or the teacher models used implicitly or explicitly in the synthesis of privileged information, contain inherent biases, DCPI could inadvertently concentrate or even reshape these biases within the reduced dataset. Therefore, rigorous auditing of both the source data and the characteristics of the synthesized privileged information will be essential to ensure that the benefits of dataset condensation do not come at the cost of fairness or equity in downstream applications.


\end{document}